\documentclass{article}

\usepackage[main, final]{neurips_2026}

\usepackage[utf8]{inputenc} 
\usepackage[T1]{fontenc}    
\usepackage{hyperref}       
\usepackage{url}            
\usepackage{booktabs}       
\usepackage{amsfonts}       
\usepackage{nicefrac}       
\usepackage{microtype}      
\usepackage{xcolor}         
\usepackage{amsmath,bm}
\usepackage{amssymb}
\usepackage{wrapfig}
\usepackage{graphicx}
\usepackage{subcaption}
\usepackage{caption} 
\usepackage{booktabs}
\usepackage{placeins}
\usepackage{multirow}
\usepackage{needspace}
\usepackage{algorithm}
\usepackage{algpseudocode}
\usepackage{pifont}
\usepackage{titletoc}
\usepackage{hyperref}
\usepackage{authblk}

\newcommand{\xmark}{\ding{55}}  

\title{Winfree Oscillatory Neural Network}

\author[1,2]{\textbf{Jiawen~Dai}}
\author[1]{\textbf{Yue~Song}$^\dagger$}

\affil[1]{College of AI, Tsinghua University}
\affil[2]{Shanghai Jiao Tong University}

\begin{document}

\maketitle

\begingroup
\renewcommand\thefootnote{$^\dagger$}
\footnotetext{Correspondence to: yue-song@mail.tsinghua.edu.cn}
\endgroup

\begin{abstract}
Oscillations and synchronization are widely believed to play a fundamental role in representation and computation. However, existing machine learning approaches based on synchronization dynamics have largely been confined to specialized settings such as object discovery, with limited evidence of scalability to standard vision benchmarks or logic reasoning tasks. We propose the \emph{Winfree Oscillatory Neural Network} (\emph{WONN}), a dynamical neural architecture based on generalized Winfree dynamics. \emph{WONN} evolves representations on the torus $(S^1)^d$ through structured oscillatory interactions, combining phase-based inductive biases with flexible and hierarchical interaction mechanisms instantiated as either fixed trigonometric mappings or learnable neural networks. We evaluate \emph{WONN} on image recognition and complex reasoning tasks, including CIFAR, ImageNet, Maze-hard, and Sudoku. Across these domains, \emph{WONN} achieves competitive or superior performance with strong parameter efficiency. \textbf{In particular, \emph{WONN} is, to our knowledge, the first synchronization-based oscillatory architecture to scale competitively to ImageNet-1K.} Furthermore, on Maze-hard, \emph{WONN} achieves \textbf{80.1\% accuracy using only 1\% of the parameters of prior state-of-the-art models}. These results suggest that structured oscillatory dynamics provide a scalable and parameter-efficient alternative to conventional neural architectures. The project page is available at \url{https://jiawen-dai.github.io/WONN_Project_Page/}.
\end{abstract}

\Needspace{11\baselineskip}
\begin{wrapfigure}{r}{0.5\columnwidth}
    \vspace{-3.5\baselineskip}
    \centering
    \includegraphics[width=\linewidth]{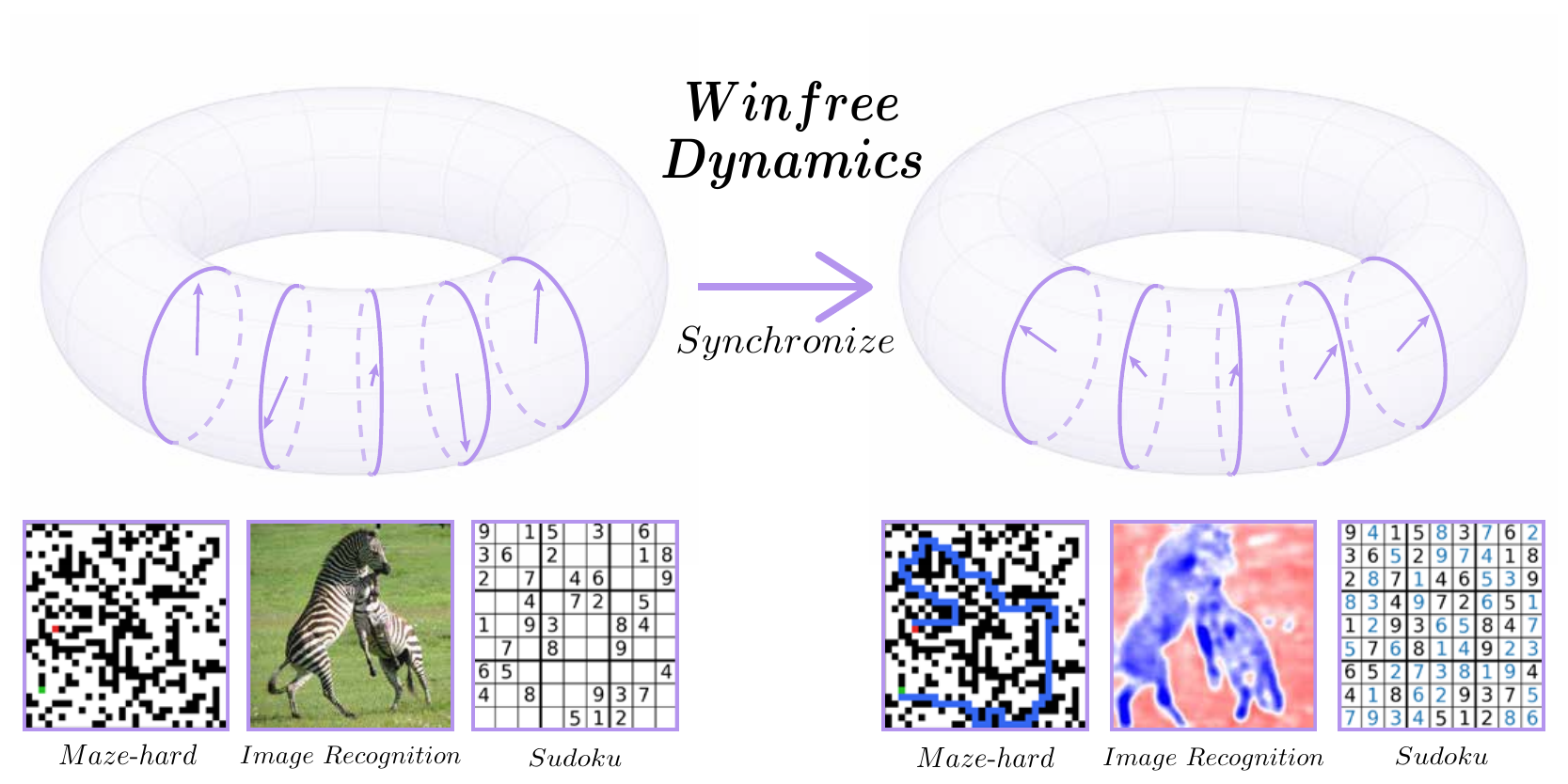}
    \caption{
    Illustration of the proposed \emph{WONN}. The network evolves on the toroidal phase space $(S^1)^d$, where neurons are represented as phase oscillators. Through generalized Winfree synchronization dynamics, oscillators self-organize into structured collective states, enabling effective computation for both image recognition and reasoning tasks.
    }
    \label{fig:teaser}
    \vspace{-0.8\baselineskip}
\end{wrapfigure}

\section{Introduction}

Oscillations and synchronization are ubiquitous mechanisms for organizing collective dynamics in biological and physical systems~\citep{buzsaki2006rhythms,winfree1967biological,kuramoto1984chemical}. In neuroscience, oscillatory activity has long been associated with temporal coordination, inter-area communication, and structured information processing~\citep{buzsaki2006rhythms,fries2005mechanism,muller2018cortical}, while synchronization and temporal correlation have been proposed as mechanisms for perceptual binding and coherent representation formation~\citep{treisman1980feature,vondermalsburg1981correlation,singer1995visual}. In physics, coupled oscillator systems provide canonical examples of how simple local interactions can self-organize into coherent macroscopic behavior~\citep{winfree1967biological,kuramoto1984chemical}. Together, these observations suggest an alternative perspective on neural computation: rather than viewing representation learning purely as a sequence of static feature transformations, computation may instead emerge from the collective evolution and synchronization of interacting dynamical states.

Despite their conceptual appeal, synchronization-based principles have so far had limited impact on mainstream machine learning. Existing synchronization-inspired approaches are primarily restricted to specialized settings such as object-centric representation learning or unsupervised object discovery~\citep{lowe2022cae,stanic2023ctcae,gopalakrishnan2024syncx,lowe2023rotating}. More recent architectures such as \emph{AKOrN}~\citep{miyato2025akorn} introduce Kuramoto-type oscillatory neurons into neural networks, but the empirical scalability of oscillatory architectures remains largely unclear. In particular, it is unknown whether synchronization-based computation can scale to large-scale image recognition benchmarks such as ImageNet-1k~\citep{deng2009imagenet}, or whether oscillatory dynamics can provide effective inductive biases for difficult combinatorial reasoning tasks such as maze pathfinding~\citep{lehnert2024beyond,wang2025hrm}. This raises a fundamental question: \emph{Can synchronization-driven dynamics serve as a scalable and parameter-efficient alternative to conventional feed-forward or transformer-style neural architectures?}

In this work, we introduce the \emph{Winfree Oscillatory Neural Network} (\emph{WONN}), a neural architecture built upon generalized Winfree synchronization dynamics~\citep{winfree1967biological,manoranjani2023generalization}. As illustrated in Fig.~\ref{fig:teaser}, \emph{WONN} evolves neural representations on a high-dimensional toroidal phase space $(S^1)^d$, where each neuron is represented as a phase oscillator. Unlike Kuramoto-type models based primarily on pairwise phase differences, Winfree dynamics decompose interactions into separable sensitivity and influence functions, enabling more flexible synchronization behaviors and interaction structures.

\emph{WONN} combines geometric inductive biases with flexible synchronization dynamics. The interaction functions can either be instantiated as fixed trigonometric mappings or parameterized by learnable neural networks, while grouped dynamics introduce hierarchical interactions that support both local coordination and global information flow. Under trigonometric interactions, the oscillatory core further admits an interaction energy that provides a useful stability perspective and diagnostic signal for reasoning tasks. Moreover, the dual phase--frequency state design separates fast synchronization dynamics from slower frequency evolution, enabling stable iterative refinement of representations.

We evaluate \emph{WONN} on both visual perception and structured reasoning tasks. Across CIFAR and ImageNet benchmarks, \emph{WONN} achieves competitive or superior accuracy compared with convolutional, transformer-based, and synchronization-inspired baselines while using substantially fewer parameters. \textbf{Importantly, \emph{WONN} is, to our knowledge, the first synchronization-based oscillatory architecture to scale to ImageNet-1K with competitive performance against modern architectures such as ResNet and ViT.} The learned representations further exhibit characteristic bimodal phase organization, where distinct synchronized phase groups specialize to complementary coarse and fine image structures. On complex reasoning tasks, \textbf{\emph{WONN} achieves competitive Maze-hard performance using only \textbf{1\%} of the parameters of the state-of-the-art \emph{HRM}~\cite{wang2025hrm}}, while also reaching perfect test accuracy on Sudoku with strong parameter efficiency. Taken together, these results suggest that generalized Winfree synchronization dynamics provide a scalable, high-performance, and parameter-efficient computational principle for both visual perception and logical reasoning.

\section{Winfree Oscillatory Neural Networks}

This section develops \emph{WONN} by connecting classical Winfree dynamics with learnable neural architectures. We first review the underlying Winfree model and its generalizations (Sec.~\ref{sec:winfree_model}), then introduce the \emph{WONN} architecture with structured interactions (Sec.~\ref{sec:winfree_network}), and finally discuss its key dynamical and representational properties (Sec.~\ref{sec:winfree_property}).

\subsection{From Winfree Dynamics to Learnable Oscillatory Systems}
\label{sec:winfree_model}
\noindent\textbf{Winfree Dynamics.}
We consider a population of coupled phase oscillators that are governed by the Winfree model \cite{winfree1967biological}:
\begin{equation}
\dot{\theta}_i = \omega_i + \frac{\kappa}{N} S(\theta_i)\sum_{j=1}^{N} I(\theta_j), \quad i=1,\dots,N,
\label{eq:winfree}
\end{equation}

where $\theta_i \in \mathbb{T} := \mathbb{R}/(2\pi\mathbb{Z})$ is the phase of the $i$-th oscillator, $\omega_i \in \mathbb{R}$ is its natural frequency and $\kappa \in \mathbb{R}_+$ is the coupling strength. The interaction is mediated through a mean-field aggregation of the influence function $I(\cdot)$, modulated by the sensitivity function $S(\cdot)$.

\noindent\textbf{Structured Interactions Beyond Kuramoto.}
A widely studied special case of oscillator synchronization is the Kuramoto model, whose dynamics are governed by pairwise phase differences:
\begin{equation}
    \dot{\theta}_i
    =
    \omega_i
    +
    \gamma \sum_{j=1}^{N} K_{ij}\sin(\theta_j-\theta_i),
    \quad t>0,
\label{eq:kuramoto_model} 
\end{equation}
where $\gamma \in \mathbb{R}_+$ denotes the coupling strength. These dynamics preserve global phase-shift symmetry under $\theta_i \mapsto \theta_i+\alpha$, since the interaction depends only on relative phase differences. While this symmetry simplifies analysis, it may also limit the diversity of attainable collective dynamics.

By contrast, the Winfree formulation allows more general interaction structures and naturally extends beyond this symmetric regime. In particular, introducing additional coupling terms yields
\begin{equation}
\dot{\theta}_i = \omega_i + \gamma \sum_{j=1}^{N} K_{ij}\big[\sin(\theta_j - \theta_i) + q \sin(\theta_j + \theta_i)\big],
\label{eq:generalized_winfree}
\end{equation}
where $q$ controls the degree of symmetry breaking. This formulation recovers the Kuramoto model when $q=0$, and produces asymmetric interactions for $q \neq 0$, enabling richer synchronization patterns. When $q=1$, the dynamics can be rewritten as $\dot{\theta}_i = \omega_i + 2\gamma \cos\theta_i \sum_{j=1}^{N} K_{ij} \sin(\theta_j)$, which corresponds to a separable Winfree form with $S(\theta)=\cos(\theta)$ and $I(\theta)=\sin(\theta)$, arising from the first-order Fourier components of the interaction functions~\cite{manoranjani2023generalization}.

These formulations highlight a key perspective: \emph{collective dynamics are determined by the structure of interactions rather than pairwise phase differences alone}. This flexibility makes Winfree-type systems a natural candidate for learnable neural architectures.

\noindent\textbf{Geometric structure.} Each oscillator evolves on the circle $\mathbb{T} := \mathbb{R}/(2\pi\mathbb{Z}) \cong S^1$, and the full system resides on the torus
\begin{equation}
\bm{\Theta} = (\theta_1,\dots,\theta_d) \in \mathcal{M} := \mathbb{T}^d \cong (S^1)^d.
\end{equation}
The dynamics define a vector field on $\mathcal{M}$, while updates can be locally represented in the tangent space $T_{\bm{\Theta}}\mathcal{M} \cong \mathbb{R}^d$. In practice, this suggests performing computations in a Euclidean embedding while enforcing periodic structure in the phase variables.

\subsection{Winfree Oscillatory Neural Network}
\label{sec:winfree_network}

\begin{figure}[t]
    \centering
    \includegraphics[width=0.8\columnwidth]{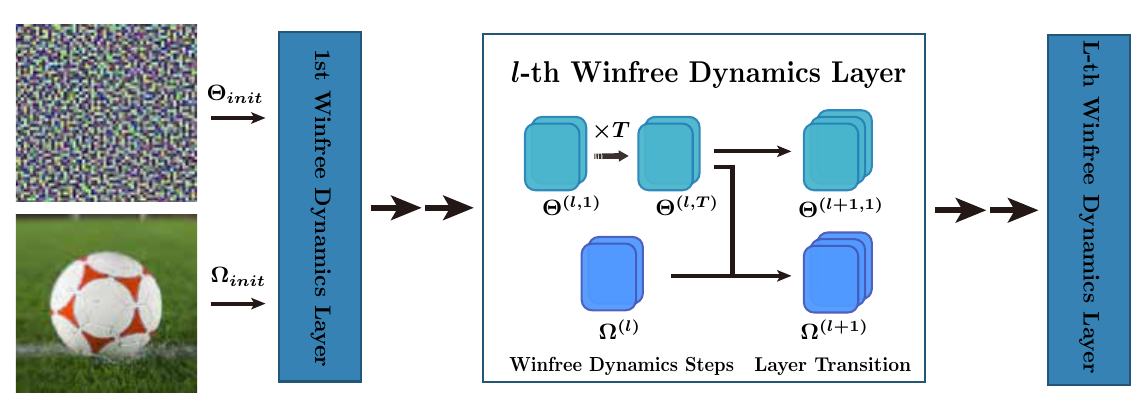}
    \caption{Overview of the \emph{WONN} architecture. The input is encoded into an initial frequency state $\bm{\Omega_{{init}}}$, while the phase state $\bm{\Theta_{init}}$ is initialized randomly. Computation proceeds through stacked Winfree dynamics layers, each consisting of multiple parameter-shared recurrent dynamics steps followed by layer-transition updates of both phase and frequency states. Through this iterative synchronization process, \emph{WONN} evolves structured oscillatory representations across layers on the toroidal phase space before producing the final prediction.
    }
    \label{fig:net}
\end{figure}

Fig.~\ref{fig:net} illustrates the overall architecture of \emph{WONN}, where representations are computed through iterative synchronization dynamics between coupled oscillatory neurons.

\noindent \textbf {State and initialization.} Each layer maintains a dual state consisting of a phase variable $\bm{\Theta} \in (S^1)^d$ and a frequency variable $\bm{\Omega} \in \mathbb{R}^d$. The phase captures oscillatory structure, while the frequency carries input-dependent information. Given an input $x$, we initialize $\bm{\Omega}^{(0)} = \bm{\Omega_{init}} = f_{\mathrm{init}}(x)$ and $\bm{\Theta}^{(0)}  = \bm{\Theta_{init}}  \sim \mathcal{N}(0, \sigma^2)$, where $f_{\mathrm{init}}$ is a learnable embedding.

\noindent \textbf{Discrete Winfree dynamics.} At the core of \emph{WONN} is a discretized Winfree evolution. For each layer $l$, we perform $T$ recurrent updates:
\begin{equation}
\theta_i^{(l,t+1)} = \theta_i^{(l,t)} + \gamma \Big[
\omega_i^{(l)} + S(\theta_i^{(l,t)}) \sum_j c_{ij} I(\theta_j^{(l,t)})
\Big], \quad t=1,\dots,T.
\label{eq:discrete_winfree}
\end{equation}
with parameters shared across time steps. Here $c_{ij}$ denotes the coupling coefficient between oscillators $i$ and $j$. The above ODE defines a controlled dynamical system over phases.

\noindent \textbf{Interaction mechanisms.} The behavior of the system is governed by the parameterization of the sensitivity and influence functions $S$ and $I$. In the simplest case, these are defined point-wise as $S(\theta_i)$ and $I(\theta_i)$, implemented either using fixed trigonometric functions or learnable neural mappings.

\Needspace{14\baselineskip}
\begin{wrapfigure}[14]{r}{0.33\columnwidth}
    \vspace{-1.8\baselineskip}
    \centering
    \includegraphics[width=\linewidth]{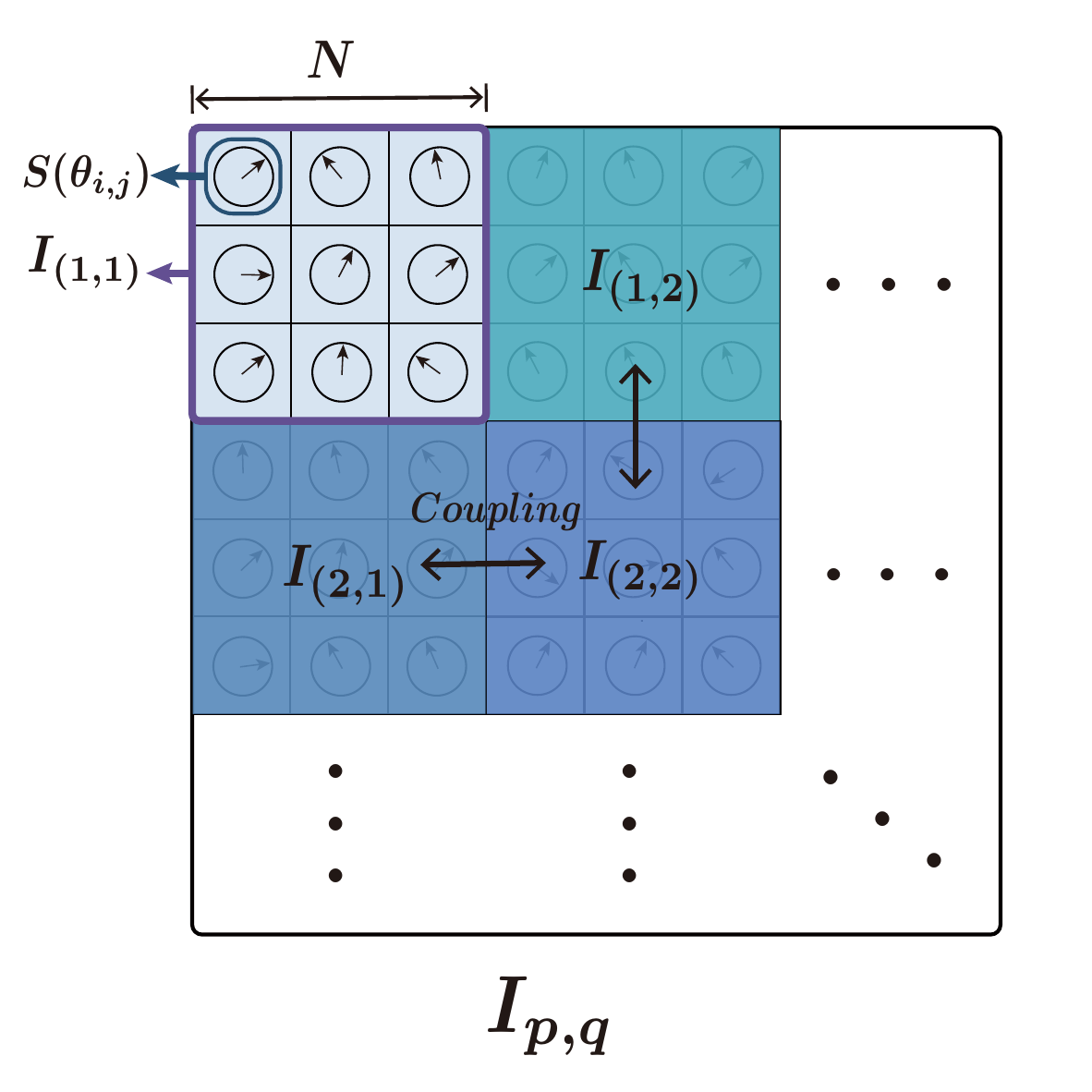}
    \caption{Illustration of $S$ and $I$.}
    \label{fig:grouped_winfree}
    \vspace{-0.8\baselineskip}
\end{wrapfigure}

To capture structured interactions, we introduce a \textit{\textbf{grouped formulation}}. As shown in Fig.~\ref{fig:grouped_winfree}, oscillators are partitioned into spatial patches $\mathcal{G}_{p,q}$, where each patch aggregates local states into a shared influence signal:
\begin{equation}
I^{\text{patch}}_{p,q} = I_{\text{patch}}(\{\theta_{i,j}\}_{(i,j)\in \mathcal{G}_{p,q}}).
\label{eq:grouped_winfree} 
\end{equation}
Here $\mathcal{G}_{p,q}$ denotes an $N \times N$ region, where $N$ is the group size. This induces a hierarchical interaction pattern, combining intra-group synchronization with inter-group coordination.

More generally, the coupling coefficients $\sum c_{ij}$ can be implemented through local or global components. Local interactions are implemented via convolutional operators with bounded receptive fields, while global interactions are realized through attention mechanisms. This results in a heterogeneous, state-dependent mean field, extending the homogeneous coupling in classical Winfree dynamics. We use global attentive coupling for all \emph{WONN} models throughout the paper, and discuss local convolutional coupling as an alternative in Appendix~\ref{sec:additional_image_recognition_results_and_ablations}.

\begin{equation}
\mathcal{G}_{p,q} =
\left\{
(i,j)\in \mathbb{Z}^2
\;\middle|\;
(p-1)N+1 \le i \le pN,\;
(q-1)N+1 \le j \le qN
\right\}.
\label{eq:patch-index}
\end{equation}

\begin{algorithm}[h]
\caption{Forward pass of \emph{WONN}}
\label{alg:wonn_train}
\begin{algorithmic}[1]
\State \textbf{Input:} input $x$, layers $L$, steps $\{T_l\}_{l=1}^{L}$, step size $\gamma$, group size $N$
\State \textbf{Output:} model output $y$
\State $\Omega^{(1)} \leftarrow f_{\mathrm{init}}(x)$, $\Theta^{(1,1)} \sim \mathcal{N}(0,\sigma^2 I)$
\For{$l=1,\ldots,L$}
    \For{$t=1,\ldots,T_l$}
        \State \textbf{(1) Point-wise Sensitivity Function}
        \State $S^{(l,t)} \leftarrow S_l(\Theta^{(l,t)})$
        \State \textbf{(2) Patch-wise Influence Function}
        \State partition $\Theta^{(l,t)}$ into non-overlapping $N\times N$ groups $\{\mathcal{G}_{p,q}\}$
        \State $I^{(l,t)}_{p,q} \leftarrow I^{l}_{\mathrm{patch}}\!\left(\{\theta^{(l,t)}_{i,j}\}_{(i,j)\in \mathcal{G}_{p,q}}\right)$
        
        
        \State \textbf{(3) Winfree Dynamics Step}

        \State $M_{p,q}^{(l,t)}
        \leftarrow \sum_j c_{ij} I_j^{(l,t)}$
        
        \State $\widetilde{\mathbf{M}}^{(l,t)}
        \leftarrow \texttt{Broadcast}\!\left(
        \mathbf{M}_{p,q}^{(l,t)}\right)$
        \Comment{repeat within each group}
        
        \State $\Delta\Theta^{(l,t)}
        \leftarrow \Omega^{(l)}
        + S^{(l,t)} \odot \widetilde{\mathbf{M}}^{(l,t)}$
        
        \State $\Theta^{(l,t+1)}
        \leftarrow \mathrm{wrap}_{[-\pi,\pi)}\!\left(
        \Theta^{(l,t)}+\gamma\Delta\Theta^{(l,t)}
        \right)$
    \EndFor
        \State \textbf{(4) Layer Transition}
        \State $\Theta^{(l+1,1)} \leftarrow \texttt{Theta-Update}_l(\Theta^{(l,T_l)})$,\quad $\Omega^{(l+1)} \leftarrow \texttt{Omega-Update}_l(\Omega^{(l)},\Theta^{(l,T_l)})$
\EndFor
\State \textbf{(5) Output}
\State $y \leftarrow \texttt{OutputHead}(\Theta^{(L,T_L)})$
\State \textbf{return} $y$
\end{algorithmic}
\end{algorithm}

\noindent \textbf{Layer transitions.} Within each layer, $\Omega^{(l)}$ remains fixed during the
$T$ phase updates in Eq.~(5). After these updates, the phase
and frequency are jointly updated as
\begin{align}
\bm{\Theta}^{(l+1)} = \texttt{Theta-Update}(\bm{\Theta}^{(l,T)}),\quad
\bm{\Omega}^{(l+1)} = \texttt{Omega-Update}(\bm{\Omega}^{(l)}, \bm{\Theta}^{(l,T)}).
\end{align}
This design separates fast phase evolution within each layer from slower updates of the frequency state across layers, allowing oscillatory patterns to be progressively refined.

Since phases lie on the circle, computations are performed via Euclidean embeddings while preserving periodic structure. Each phase is represented as $(\sin\theta, \cos\theta)$, and the phase update $\texttt{Theta-Update}$ first applies weight-tied convolutional operators in this embedding space (which can be interpreted as linear operators acting on complex representation $e^{i\bm{\Theta}}$) and maps the result back to the torus via
\begin{align}
\bm{\Theta}_{\text{new}} = \texttt{atan2}\big(\texttt{Conv}(\sin\bm{\Theta}), \texttt{Conv}(\cos\bm{\Theta})\big)
\end{align}
For frequency updates, we similarly construct a Euclidean
embedding of $\Theta$ and combine it with the current
frequency state:
\begin{equation}
\begin{aligned}
\texttt{ThetaEmbedding}^{(l)}(\Theta)
&= \texttt{Conv}_{\Theta}^{(l)}
\left(\texttt{Concat}[\sin\Theta,\cos\Theta]\right),\\
\Omega^{(l+1)}
&= \texttt{Conv}_{\Omega}^{(l)}
\left(\texttt{Concat}\left[
\texttt{ThetaEmbedding}^{(l)}(\Theta^{(l,T)}),
\Omega^{(l)}
\right]\right).
\end{aligned}
\end{equation}
Here, $\texttt{Conv}_{\Omega}^{(l)}$ denotes the learnable
convolutional fusion block, including normalization
and activation. The final model representation is obtained from the phase variables $\bm{\Theta}^{(L)}$ using a task-specific neural network head. Alg.~\ref{alg:wonn_train} summarizes a single forward pass of our \emph{WONN}.

\subsection{Dynamical and Structural Properties of \emph{WONN}}
\label{sec:winfree_property}

\emph{WONN} exhibits several distinctive properties that emerge from its oscillatory dynamics and structured interaction design. We highlight the key properties below.

\noindent\textbf{Oscillatory inductive bias.} \emph{WONN} imposes a geometric inductive bias by evolving representations on the torus $(S^1)^d$, enforcing periodic structure in the representation space. This encourages phase-based organization of features, where information can be encoded through relative phase configurations. Empirically, we observe that oscillators tend to form structured phase distributions, often exhibiting multi-modal patterns (\emph{e.g.,} phase bifurcation with modes separated by $\pi$). Such phase organization may correlate with meaningful feature extraction. For instance in vision tasks, we observe that different phase modes align with distinct components such as object edges (see Sec.~\ref{sec:image_recognition}).

\noindent\textbf{Hierarchical interaction structure.} The grouped Winfree formulation introduces a two-level interaction hierarchy. Within each group, oscillators synchronize through shared local aggregation, while interactions across groups enable global coordination. \textit{This structure interpolates between purely local and fully global coupling, controlled by the group size $N$.} In the limiting case $N{=}1$, the model reduces to point-wise interactions, whereas larger groups introduce shared contextual influence. This hierarchical design allows \emph{WONN} to adapt its interaction scale to different tasks, balancing locality and global coherence.

\noindent\textbf{Flexible interaction parameterization.}
\emph{WONN} allows flexible choices of the sensitivity and influence functions $S$ and $I$. In particular, these functions can be instantiated either as fixed trigonometric mappings (\emph{e.g.,} $S(\theta){=}\cos\theta$, $I(\theta){=}\sin\theta$) or as learnable neural networks (\emph{e.g.,} MLPs). The trigonometric parameterization provides a structured and stable interaction form aligned with classical oscillator models, while the parameterization of learnable neural networks increases expressivity. This flexibility allows to trade off inductive bias and representational capacity across different tasks.



\noindent\textbf{Energy structure and stability.}
The phase dynamics of \emph{WONN} inherit stability properties from the underlying Winfree system. Due to the periodic geometry of $(S^1)^d$, the natural frequency term cannot, in general, be associated with a globally defined potential (see Appendix~\ref{app:lyapunov}). However, when the natural frequency is excluded, the coupling is fixed and symmetric, and the interaction functions take separable trigonometric forms, the dynamics admit a Lyapunov function that decreases along trajectories, leading to stable synchronization patterns and attractors in phase space. In this regime, the interaction energy can be written as $E_{\mathrm{int}} = - \sum_{i} \sin \theta_i \sum_{j} c_{ij} \sin(\theta_j)$, capturing phase alignment under the coupling coefficients. Although this continuous-time analysis does not provide a general convergence guarantee for the full learned \emph{WONN}, the interaction energy serves as a practical diagnostic: in Maze reasoning tasks (Sec.~\ref{sec:Maze_hard}), we use it as a proxy for solution quality and perform voting to improve performance.

\noindent\textbf{Separation of timescales.}
\emph{WONN} separates fast oscillatory dynamics from slower feature evolution through its dual-state design. Within each layer, the phase state $\bm{\Theta}$ undergoes multiple recurrent updates, enabling rapid synchronization and local reorganization on $(S^1)^d$. In contrast, the frequency state $\bm{\Omega}$ is updated once per layer, acting as a slower carrier of information across layers. This separation of timescales decouples transient oscillatory computation from cross-layer feature transformation, which is reminiscent of hierarchical processing in biological systems~\cite{hasson2008hierarchy,honey2012slow,golesorkhi2021temporal,wang2025hrm}.

\section{Experiments}
\label{sec:experiments}

\subsection{Experimental Setup}

We evaluate \emph{WONN} on a diverse set of benchmarks spanning visual perception and logical reasoning. The visual tasks include CIFAR-10/100~\cite{krizhevsky2009learning} and ImageNet-100/1K~\cite{deng2009imagenet}, while the reasoning tasks consist of Maze-hard pathfinding~\citep{lehnert2024beyond,wang2025hrm} and Sudoku solving~\cite{wang2019satnet}.

\FloatBarrier
\noindent
\begin{minipage}[t]{\columnwidth}
    \centering
    \captionsetup{hypcap=false}
    \captionof{table}{Image classification results on \textbf{CIFAR-10/100}. $\dagger$ ViT models are trained with additional regularization (e.g., CutMix and label smoothing), while other models follow a plain training protocol.}
    \label{tab:cifar_results}
    \small
    \setlength{\tabcolsep}{4pt}
    \begin{tabular}{l|cc|cc}
        \toprule
        \multirow{2}{*}{\textbf{Models}}
        & \multicolumn{2}{c|}{\textbf{CIFAR-10}}
        & \multicolumn{2}{c}{\textbf{CIFAR-100}} \\
        \cmidrule(lr){2-3} \cmidrule(lr){4-5}
        & \textbf{Accuracy (\%)} & \textbf{$\#$ Parameters} & \textbf{Accuracy (\%)} & \textbf{$\#$ Parameters} \\
        \midrule
        \textbf{ResNet-18}     & 93.48 {\tiny$\pm 0.16$} & 11.17M & 70.53 {\tiny$\pm 0.10$}  & 11.22M \\
        \textbf{ResNet-50}     & 94.22 {\tiny$\pm 0.14$} & 23.52M & 73.54 {\tiny$\pm 0.41$} & 23.71M \\
        \textbf{ViT-T}$^{\dagger}$          & 90.34 {\tiny$\pm 0.18$} & 5.36M & 67.59 {\tiny$\pm 0.44$} & 5.38M \\
        \textbf{ViT-S}$^{\dagger}$          & 93.43 {\tiny$\pm 0.29$} & 21.34M & 71.18 {\tiny$\pm 0.62$}  & 21.38M \\
        \textbf{ViT-B}$^{\dagger}$         & 92.04 {\tiny$\pm 0.25$} & 85.15M & 71.05 {\tiny$\pm 0.40$} & 85.22M \\
        \emph{\textbf{AKOrN}}$^{attn}$  & 93.66 {\tiny$\pm$ 0.17} & 4.60M & 72.03 {\tiny$\pm$ 0.34} & 4.62M \\
        \midrule
        \multicolumn{5}{l}{\textit{$S(\theta_i)\&I(\theta_i)$ as MLPs}}\\
        \emph{\textbf{WONN}} ($\mathrm{Ch}=128 \to 128$) & 94.55 {\tiny$\pm$ 0.09} & 3.08M & 73.77 {\tiny$\pm$ 0.40} & 3.09M \\
        \emph{\textbf{WONN}} ($\mathrm{Ch}=64 \to 256$) & 95.12 {\tiny$\pm$ 0.04} & 7.54M & 75.12 {\tiny$\pm$ 0.49} & 7.56M \\
        \emph{\textbf{WONN}} ($\mathrm{Ch}=256 \to 256$) & \textbf{95.24} {\tiny$\pm$ 0.12} & 12.02M & \textbf{76.20} {\tiny$\pm$ 0.45} & 12.04M \\
        \midrule
        \multicolumn{5}{l}{\textit{$S(\theta_i)\&I(\theta_i)$ as trigonometric functions}}\\
        \emph{\textbf{WONN}} ($\mathrm{Ch}=128 \to 128$) & 94.50 {\tiny$\pm 0.15$} & 2.98M & 74.48 {\tiny$\pm 0.16$} & 3.00M \\
        \emph{\textbf{WONN}} ($\mathrm{Ch}=64 \to 256$) & 95.08 {\tiny$\pm 0.07$} & 7.40M & 75.81 {\tiny$\pm 0.33$}  & 7.43M \\
        \emph{\textbf{WONN}} ($\mathrm{Ch}=256 \to 256$) & \textbf{95.26} {\tiny$\pm 0.05$} & 11.84M & \textbf{76.17} {\tiny$\pm 0.52$} & 11.86M \\
        \bottomrule
    \end{tabular}
\end{minipage}
\FloatBarrier

\noindent\textbf{Visual Perception.} For image classification, we benchmark \emph{WONN} against representative convolutional and transformer-based architectures, including \emph{ResNet-18} and \emph{ResNet-50}~\cite{he2016deep}, as well as Vision Transformers (\emph{ViT-Tiny/Small/Base})~\cite{dosovitskiy2021image,touvron2021training}. We also include \emph{AKOrN}~\cite{miyato2025akorn}, a recent synchrony-based model, as a closely related baseline. \textbf{\textit{All methods are trained under standard, architecture-appropriate protocols, and no additional training techniques are introduced to favor WONN, ensuring a fair comparison.}}

\noindent\textbf{Logical Reasoning.} For reasoning tasks, we evaluate \emph{WONN} on Maze-hard and Sudoku. On Maze-hard, we mainly compare against both general-purpose LLMs (\emph{e.g.,} \emph{DeepSeek-R1}~\cite{deepseekai2025deepseekr1}) and specialized recurrent reasoning architectures, including \emph{HRM}~\cite{wang2025hrm} and \emph{TRM}~\cite{jolicoeurmartineau2025trm}. On Sudoku, we compare with \emph{AKOrN}~\cite{miyato2025akorn} and several other neural reasoning baselines~\cite{wang2019satnet,palm2018recurrent,yang2023recurrent}.

Other implementation details and extended experimental results are referred to in Appendix~\ref{sec:implementation_results_app}.

\subsection{Image Classification on CIFAR and ImageNet}
\label{sec:image_recognition}

We evaluate \emph{WONN} on standard image classification benchmarks, including CIFAR-10/100 and ImageNet-100/1K. We consider two parameterizations of the interaction functions $S$ and $I$: (i) learnable MLP-based mappings and (ii) fixed trigonometric mappings. All models use grouped Winfree dynamics in Eq.~\eqref{eq:grouped_winfree} with group size $N=2$. To assess the scalability of model size, we study three model sizes with approximately 3M, 7.5M, and 12M parameters.

\noindent\textbf{Results on CIFAR-10/100.} Table~\ref{tab:cifar_results} compares \emph{WONN} with \emph{ResNet}, \emph{ViT}, and \emph{AKOrN}, with results averaged over three random seeds. \emph{WONN} consistently outperforms all baselines on both CIFAR-10 and CIFAR-100 while using substantially fewer parameters. Notably, \emph{WONN}, trained with a plain protocol, surpasses transformer-based baselines that rely on additional regularization. These results demonstrate the effectiveness and parameter efficiency of oscillatory dynamics.

\noindent\textbf{Results on ImageNet-100/1K.}
Table~\ref{tab:imagenet_results} shows that the trends observed on CIFAR persist at larger scale: \emph{WONN} achieves competitive accuracy with substantially fewer parameters than standard baselines. In particular, \emph{WONN} achieves comparable ImageNet-1K accuracy to \emph{ResNet-50} while using approximately half the parameters (12.28M vs.\ 25.56M), and outperforms the evaluated \emph{ViT} baselines. \textbf{\textit{To the best of our knowledge, this is the first synchrony-based architecture that scales effectively to large-scale image classification.}}

\begin{figure}[t]
    \centering
    \includegraphics[width=\columnwidth]{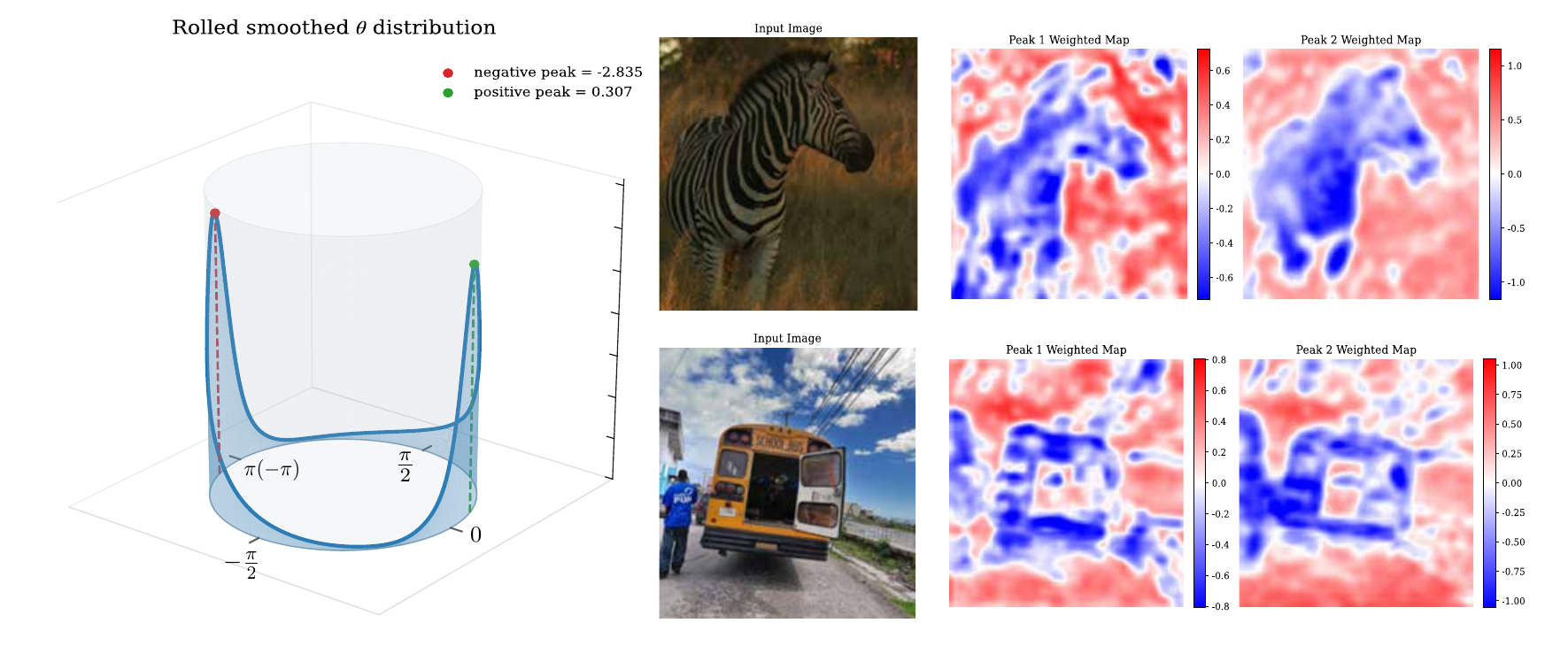}
    \caption{Visualization of the final phase distribution in \emph{WONN} on image recognition tasks. \textbf{Left:} the phase variables $\theta$ exhibit a characteristic bimodal distribution, with two dominant peaks approximately separated by $\pi$. \textbf{Right:} heatmaps constructed by weighting oscillators according to their proximity to each phase peak. The two modes capture complementary object structures: one emphasizes coarse global regions, while the other focuses on finer local details and boundaries. Together, the synchronized phase organization produces structured object-aware representations.
    }
    \label{fig:two_peak}
\end{figure}

\FloatBarrier
\noindent
\begin{minipage}{\columnwidth}
    \centering
    \captionsetup{hypcap=false}
    \captionof{table}{Image classification results on \textbf{ImageNet-100} and \textbf{ImageNet-1K}. $\dagger$ViT models are trained with additional regularization techniques compared to other models. Underlining denotes our reimplementation.}
    \label{tab:imagenet_results}
    \small
    \setlength{\tabcolsep}{4pt}
    \begin{tabular}{l|cc|cc}
        \toprule
        \multirow{2}{*}{\textbf{Models}}
        & \multicolumn{2}{c|}{\textbf{ImageNet-100}}
        & \multicolumn{2}{c}{\textbf{ImageNet-1k}} \\
        \cmidrule(lr){2-3} \cmidrule(lr){4-5}
        & \textbf{Accuracy(\%)} & \textbf{$\#$ Parameters} & \textbf{Accuracy(\%)} & \textbf{$\#$ Parameters} \\
        \midrule
        \textbf{ResNet-18}     & 78.20 & 11.23M & 69.73 & 11.69M \\
        \textbf{ResNet-50}         & 81.18 & 23.71M & \textbf{76.89} & 25.56M \\
        \textbf{ViT-S-16}$^{\dagger}$         & 77.96 & 21.70M & 75.54 & 22.05M \\
        \textbf{ViT-B-16}$^{\dagger}$        & 76.36 & 85.87M & 75.85 & 86.57M \\
        \emph{\textbf{AKOrN}}$^{attn}$  & \underline{80.08} & 4.62M & \underline{67.45} & 4.85M \\
        \midrule 
        \multicolumn{5}{l}{\textit{$S(\theta_i)\&I(\theta_i)$ as MLPs}}\\
        \emph{\textbf{WONN}} ($\mathrm{Ch}=128 \to 128$) & 81.50 & 3.09M & -- & -- \\
        \emph{\textbf{WONN}} ($\mathrm{Ch}=64 \to 256$) & 82.04 & 7.56M & 74.84 & 7.79M \\
        \emph{\textbf{WONN}} ($\mathrm{Ch}=256 \to 256$) & \textbf{82.88} & 12.05M & \textbf{76.80}{\tiny$\pm 0.14$} & 12.28M \\
        \midrule
        \multicolumn{5}{l}{\textit{$S(\theta_i)\&I(\theta_i)$ as trigonometric functions}}\\
        \emph{\textbf{WONN}} ($\mathrm{Ch}=128 \to 128$) & 81.76 & 3.00M & -- & -- \\
        \emph{\textbf{WONN}} ($\mathrm{Ch}=64 \to 256$) & \textbf{82.56} & 7.43M & -- & -- \\
        \emph{\textbf{WONN}} ($\mathrm{Ch}=256 \to 256$) & 82.22 & 11.87M & 76.36 & 12.10M \\
        \bottomrule
    \end{tabular}
\end{minipage}
\FloatBarrier

\noindent\textbf{Two-peak phase distributions.} To better understand the internal phase organization of \emph{WONN}, we analyze the final $\theta$ distribution in Fig.~\ref{fig:two_peak} left. Interestingly, the phase variables consistently bifurcate into two dominant modes, producing a characteristic bimodal distribution with peaks approximately separated by $\pi$. To visualize the functional role of these modes, we construct weighted activation maps by measuring the proximity of each oscillator phase to the corresponding peak.

Fig.~\ref{fig:two_peak} right reveals that the two synchronized phase modes capture \textit{complementary aspects of the object structure.} One phase mode primarily responds to coarse, global object regions, while the other emphasizes finer local structures and boundaries. Despite different spatial characteristics, both modes remain highly aligned with semantically relevant object regions. This suggests that the oscillatory synchronization dynamics naturally induce a structured phase decomposition of representations, where different synchronized phase groups specialize to distinct levels of spatial abstraction.

\textbf{\textit{Notably, the emergence of such bimodal phase organization is not explicitly imposed by any architecture or supervision objective, but instead arises spontaneously from the collective synchronization dynamics.}} We observe this bimodal organization across CIFAR-10, CIFAR-100, ImageNet-100, and ImageNet-1K, and hypothesize that this self-organized phase separation contributes to the strong recognition performance and parameter efficiency of \emph{WONN} by enabling coordinated yet diverse feature representations across oscillators. The mechanism underlying its emergence and whether it is preferable to other multimodal organizations remain open questions. Additional visualizations are provided in Appendix~\ref{app:Additional Visualization on Image Recognition}.

\subsection{Maze-hard Pathfinding}
\label{sec:Maze_hard}
We evaluate the reasoning capability of \emph{WONN} on the \textbf{Maze-hard} benchmark, which requires finding optimal paths in $30 \times 30$ mazes and tests long-horizon search and algorithmic reasoning. Following~\citep{wang2025hrm}, we construct $1,000$ training and $1,000$ test instances whose difficulty levels exceed $110$. For this task, \emph{WONN} uses point-wise interactions ($N=1$).

\begin{figure}[t]
    \centering
    \includegraphics[width=\columnwidth]{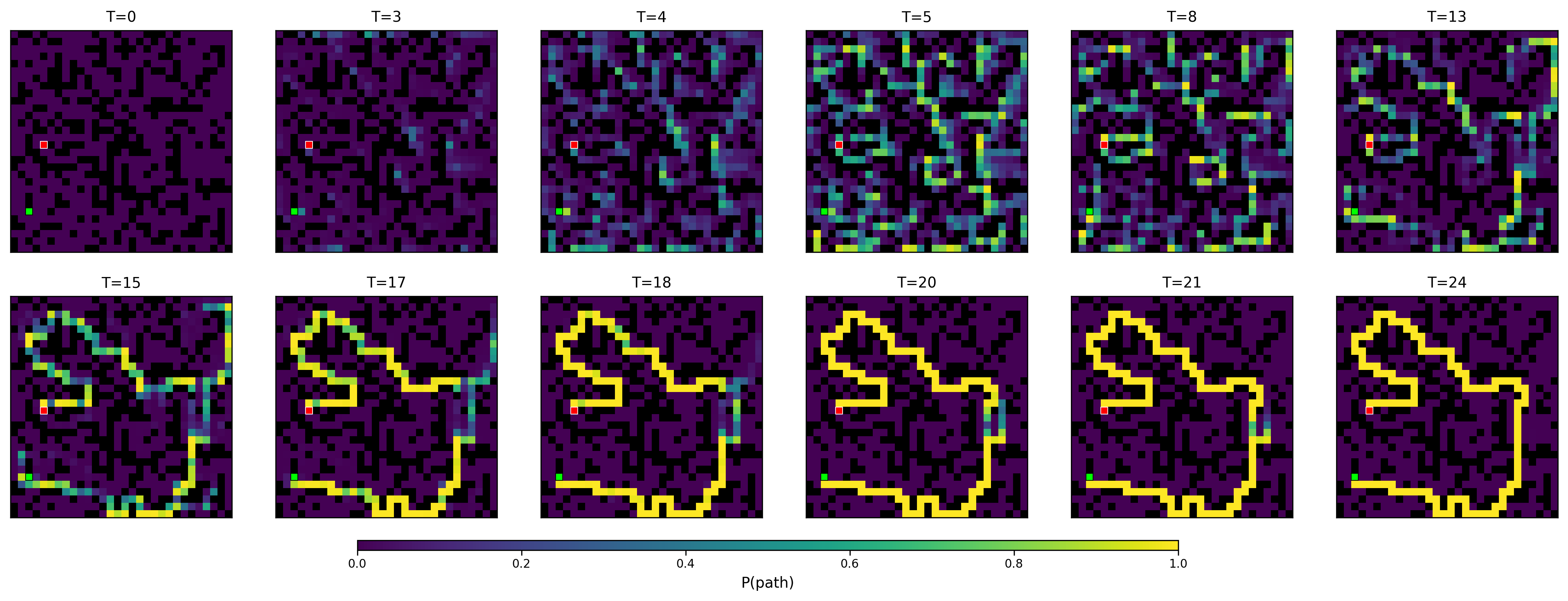}
    \caption{Synchronous path formation on Maze-hard. Each panel shows the predicted per-cell path probability. \emph{WONN} initially produces multiple diffuse path fragments. As the dynamics evolve, these fragments progressively synchronize and merge, forming a coherent path from start to goal.}
    \label{fig:maze_prob}
\end{figure}

\noindent\textbf{Results on Maze-hard.}
Table~\ref{tab:maze_results} shows that \emph{WONN} significantly outperforms LLM-based baselines and the synchrony-based model \emph{AKOrN}, and surpasses the strong recurrent model \emph{HRM}. Despite using only \textbf{0.396M} parameters (\textbf{1.47\%} of \emph{HRM}), \emph{WONN} achieves \textbf{76.2\%} accuracy. Importantly, leveraging the energy structure of the dynamics, we further improve performance via \emph{energy voting}: by sampling 32 trajectories and selecting the solution with the lowest final energy, accuracy increases to \textbf{80.1\%}, compared with 85.3\% for the 6.823M-parameter TRM-Att. This improvement highlights the practical usage of the energy formulation as a \emph{test-time selection criterion}, while the comparison with TRM-Att demonstrates a favorable accuracy--parameter-efficiency trade-off. Taken together, these results demonstrate that structured oscillatory dynamics provide an effective and highly parameter-efficient inductive bias for long-horizon reasoning.

\noindent\textbf{Synchronous Pathfinding on Maze-hard.}
To better understand the role of oscillatory dynamics, we visualize the intermediate states of \emph{WONN} on Maze-hard (Fig.~\ref{fig:maze_prob}). We plot heatmaps of the predicted per-cell path probability, revealing how solutions emerge over time.

\FloatBarrier
\noindent
\begin{minipage}{\columnwidth}
    \centering
    \captionsetup{hypcap=false}
    \captionof{table}{Maze-hard pathfinding results. \emph{Energy Voting} denotes selecting the solution with the lowest energy among $32$ independently sampled trajectories.}
    \label{tab:maze_results}
    \small
    \setlength{\tabcolsep}{4pt}
    \begin{tabular}{lcc}
        \toprule
        \textbf{Models} & \textbf{Accuracy (\%)} & \textbf{$\#$ Parameters} \\
        \midrule
        \multicolumn{3}{l}{\textit{LLMs}} \\
        \textbf{Deepseek R1}     & 0.0 & 671B \\
        \textbf{Claude 3.7 8K}   & 0.0 & ? \\
        \textbf{O3-mini-high}    & 0.0 & ? \\
        \midrule
        \multicolumn{3}{l}{\textit{Recurrent models}} \\
        \textbf{HRM}    & 74.5 & 27M \\
        \textbf{TRM-Att} (dihedral augmented)    & \textbf{85.3} & 7M \\
        \textbf{TRM-MLP} (dihedral augmented)    & 0.0 & 19M \\
        \midrule
        \multicolumn{3}{l}{\textit{Other synchrony-based model}} \\
        \textbf{\emph{AKOrN}}$^{attn}$   & 36.2 & 1M \\
        \midrule
        \multicolumn{3}{l}{\textit{$S(\theta_i)$ \& $I(\theta_i)$ as trigonometric functions}} \\
        \emph{\textbf{WONN}} (K=1)           & \textbf{76.2} & \textbf{0.396M} \\
        \emph{\textbf{WONN}} (Energy Voting, K=32)      & \textbf{80.1} & \textbf{0.396M} \\
        \bottomrule
    \end{tabular}
\end{minipage}
\FloatBarrier

Unlike conventional methods that construct paths incrementally or greedily, \emph{WONN} exhibits a distinct synchronization-driven process. In the early stage, the model generates multiple diffuse path fragments across the maze that correspond to candidate solution trajectories. These fragments are not locally consistent and may include invalid transitions. As the dynamics evolve, these candidate fragments begin to interact and synchronize. Inconsistent paths are suppressed, while compatible segments reinforce each other and progressively merge. This collective refinement process leads to the emergence of a single coherent path connecting the start and goal. In later stages, the model further refines the solution by removing redundant segments and improving path optimality. 

Additional examples (Appendix~\ref{sec:additional_visualization_maze_hard}) show that this multi-fragment-to-single-path behavior is consistent across hard instances. These observations suggest that synchronization dynamics provide a powerful inductive bias for combinatorial search: \textit{instead of committing to a single trajectory early, the model explores multiple candidates in parallel and resolves them through collective alignment.}

\subsection{Sudoku Reasoning}
\label{sec:Sudoku}

We further evaluate \emph{WONN} on Sudoku, a standard benchmark for combinatorial reasoning. We use the dataset from~\cite{wang2019satnet}, where training instances contain 31--42 given digits, and evaluation is performed on $1{,}000$ test boards drawn from the same distribution. For this task, \emph{WONN} adopts point-wise interactions ($N=1$) and follows the same training protocol as \emph{AKOrN}.

\noindent\textbf{Results on Sudoku.} Table~\ref{tab:sudoku_results} reports results averaged over five runs. \emph{WONN} achieves perfect accuracy on the test set, matching the best-performing methods. It improves upon \emph{AKOrN} and \emph{HRM} while using significantly fewer parameters, and achieves comparable performance to both recurrent and transformer-based baselines~\cite{palm2018recurrent,yang2023recurrent}. Consistent with the Maze-hard results, these findings suggest that oscillatory dynamics provide an effective inductive bias for structured reasoning tasks.

\FloatBarrier
\noindent
\begin{minipage}{\columnwidth}
    \centering
    \captionsetup{hypcap=false}
    \captionof{table}{Sudoku solving results.}
    \label{tab:sudoku_results}
    \small
    \setlength{\tabcolsep}{5pt}
    \begin{tabular}{l|cc}
        \toprule
        \textbf{Models}
        & \textbf{Accuracy (\%)}
        & \textbf{$\#$ Parameters} \\
        \midrule
        \textbf{SAT-Net}  
        & 98.3 & 0.618M \\
        \textbf{RRN}  
        & 99.8 & 0.201M \\
        \textbf{R-Transformer}  
        & \textbf{100.0} & 0.211M \\
        \textbf{Transformer}  
        & 98.6{\tiny$\pm$ 0.3} & 16.87M \\
        \emph{\textbf{HRM}}
        & 99.7{\tiny$\pm$ 0.2} & 27.28M \\
        \emph{\textbf{AKOrN}}$^{\mathrm{attn}}$($T=16$)  
        & 99.8{\tiny$\pm$ 0.1} & 2.98M \\
        \midrule
        \multicolumn{3}{l}{\textit{$S(\theta_i)\&I(\theta_i)$ as trigonometric functions}}\\
        \emph{\textbf{WONN}} ($T=16$) 
        & \textbf{100.0{\tiny$\pm$ 0.0}} & 1.58M \\
        \bottomrule
    \end{tabular}
\end{minipage}
\FloatBarrier

\subsection{Computational Cost}
\label{sec:compute_cost}

To evaluate efficiency beyond parameter count, we benchmark training latency, single-sample inference latency, and peak GPU memory on CIFAR-10, Sudoku, and Maze-hard. All measurements are conducted on a single NVIDIA H100 using the same numerical precision and timing protocol. Training latency is measured
with batch size 32.

Table~\ref{tab:compute_cost} compares \emph{WONN} with
representative feed-forward and oscillatory baselines on
CIFAR-10. As expected for a dynamical architecture,
\emph{WONN} incurs additional latency relative to ResNet-50
and ViT-S/4 while using roughly half their parameters.
Although \emph{AKOrN} uses fewer parameters, \emph{WONN}
is faster and more memory-efficient during training.
Additional measurements on Sudoku and Maze-hard, including
the cost of energy voting, are provided in
Appendix~\ref{app:compute_cost}.

\begin{table}[htbp]
    \centering
    \caption{Computational costs on CIFAR-10, measured on a single
    NVIDIA H100 with batch sizes of 32 for training and 1 for
    inference. \emph{WONN} uses trigonometric
    interactions with 256 channels.}
    \label{tab:compute_cost}
    \small
    \setlength{\tabcolsep}{4pt}
    \begin{tabular}{lccccc}
        \toprule
        \multirow{2}{*}{\textbf{Model}}
        & \multirow{2}{*}{\textbf{Params.}}
        & \multicolumn{2}{c}{\textbf{Latency (ms)}}
        & \multicolumn{2}{c}{\textbf{Peak memory (GiB)}} \\
        \cmidrule(lr){3-4} \cmidrule(lr){5-6}
        & & Train/batch & Infer/sample & Train & Infer \\
        \midrule
        ResNet-50
        & 23.52M & 44.37 & 1.71 & 0.765 & 0.125 \\
        ViT-S/4
        & 21.34M & 37.52 & 1.87 & 0.690 & 0.113 \\
        \emph{AKOrN}
        & 4.60M & 59.61 & 3.60 & 5.243 & 0.083 \\
        \midrule
        \emph{\textbf{WONN}}
        & 11.84M & 51.74 & 3.67 & 3.445 & 0.084 \\
        \bottomrule
    \end{tabular}
\end{table}
\FloatBarrier

\section{Conclusion}

We introduced the \emph{Winfree Oscillatory Neural Network} (\emph{WONN}), a neural architecture built upon generalized Winfree synchronization dynamics. Unlike conventional architectures that primarily rely on static feature transformations, \emph{WONN} performs computation through the collective evolution and synchronization of phase oscillators on a toroidal state space $(S^1)^d$. By combining flexible interaction parameterizations, hierarchical grouped synchronization dynamics, and a dual phase--frequency state design, \emph{WONN} provides a scalable framework for oscillatory neural computation.

Empirically, we show that synchronization-based architectures can scale far beyond the limited settings previously explored in oscillatory or synchronization-inspired models. On visual perception tasks, \emph{WONN} achieves competitive or superior performance on CIFAR and ImageNet benchmarks while maintaining strong parameter efficiency. In particular, \emph{WONN} successfully scales to ImageNet-1K, which, to the best of our knowledge, makes it the first synchronization-based oscillatory architecture to achieve competitive large-scale image recognition performance. The learned phase representations further exhibit characteristic bimodal phase organization, where distinct synchronized phase groups specialize to complementary coarse and fine image structures. Beyond perception, \emph{WONN} also demonstrates strong capability on structured reasoning tasks. On Maze-hard, \emph{WONN} achieves competitive performance using only a tiny fraction of the parameters of state-of-the-art recurrent reasoning models. Visualization of the temporal evolution further reveals that the network progressively assembles local candidate path fragments into coherent global solutions, suggesting that synchronization provides an effective inductive bias for complex search and reasoning problems. On Sudoku, \emph{WONN} further achieves perfect test accuracy with strong parameter efficiency.

More broadly, our results suggest that synchronization dynamics may provide a promising alternative computational paradigm for modern machine learning. We hope this work encourages further exploration of oscillatory neural computation, synchronization-driven representation learning, and dynamical systems perspectives on large-scale neural architectures.

\begin{ack}
This work was supported by the National Natural Science Foundation of China (Grant No. 62676211). Yue Song thanks Max Welling for the fruitful discussion and his insightful feedback.
\end{ack}

\bibliographystyle{plain}
\bibliography{references}

@article{treisman1980feature,
  title     = {A feature-integration theory of attention},
  author    = {Treisman, Anne M. and Gelade, Garry},
  journal   = {Cognitive Psychology},
  volume    = {12},
  number    = {1},
  pages     = {97--136},
  year      = {1980},
  publisher = {Elsevier}
}

@techreport{vondermalsburg1981correlation,
  title={The correlation theory of brain function},
  author={von der Malsburg, Christoph},
  institution={Max-Planck-Institute for Biophysical Chemistry},
  number={Internal Report 81-2},
  year={1981}
}

@article{singer1995visual,
  title={Visual feature integration and the temporal correlation hypothesis},
  author={Singer, Wolf and Gray, Charles M.},
  journal={Annual Review of Neuroscience},
  volume={18},
  pages={555--586},
  year={1995}
}

@article{lowe2022cae,
  title   = {Complex-Valued Autoencoders for Object Discovery},
  author  = {L{\"o}we, Sindy and Lippe, Phillip and Rudolph, Maja and Welling, Max},
  journal = {Transactions on Machine Learning Research},
  year    = {2022}
}

@inproceedings{stanic2023ctcae,
  title     = {Contrastive Training of Complex-Valued Autoencoders for Object Discovery},
  author    = {Stani{\'c}, Aleksandar and Gopalakrishnan, Anand and Irie, Kazuki and Schmidhuber, J{\"u}rgen},
  booktitle = {Advances in Neural Information Processing Systems},
  volume    = {36},
  year      = {2023}
}

@inproceedings{lowe2023rotating,
  title     = {Rotating Features for Object Discovery},
  author    = {L{\"o}we, Sindy and Lippe, Phillip and Locatello, Francesco and Welling, Max},
  booktitle = {Advances in Neural Information Processing Systems},
  volume    = {36},
  year      = {2023}
}

@inproceedings{gopalakrishnan2024syncx,
  title     = {Recurrent Complex-Weighted Autoencoders for Unsupervised Object Discovery},
  author    = {Gopalakrishnan, Anand and Stani{\'c}, Aleksandar and Schmidhuber, J{\"u}rgen and Mozer, Michael Curtis},
  booktitle = {Advances in Neural Information Processing Systems},
  volume    = {37},
  year      = {2024}
}

@article{liu2026krause,
  title         = {Krause Synchronization Transformers},
  author        = {Liu, Jingkun and Yue, Yisong and Welling, Max and Song, Yue},
  journal       = {arXiv preprint arXiv:2602.11534},
  year          = {2026},
  eprint        = {2602.11534},
  archivePrefix = {arXiv},
  doi           = {10.48550/arXiv.2602.11534}
}

@article{winfree1967biological,
  title     = {Biological Rhythms and the Behavior of Populations of Coupled Oscillators},
  author    = {Winfree, A. T.},
  journal   = {Journal of Theoretical Biology},
  volume    = {16},
  number    = {1},
  pages     = {15--42},
  year      = {1967},
  publisher = {Elsevier},
  doi       = {10.1016/0022-5193(67)90051-3}
}

@book{kuramoto1984chemical,
  title     = {Chemical Oscillations, Waves, and Turbulence},
  author    = {Kuramoto, Yoshiki},
  series    = {Springer Series in Synergetics},
  volume    = {19},
  year      = {1984},
  publisher = {Springer},
  address   = {Berlin, Heidelberg},
  doi       = {10.1007/978-3-642-69689-3}
}

@inproceedings{miyato2025akorn,
  title     = {Artificial Kuramoto Oscillatory Neurons},
  author    = {Miyato, Takeru and L{\"o}we, Sindy and Geiger, Andreas and Welling, Max},
  booktitle = {International Conference on Learning Representations},
  year      = {2025}
}

@inproceedings{nguyen2024kuramatognn,
  title     = {From Coupled Oscillators to Graph Neural Networks: Reducing Over-smoothing via a Kuramoto Model-based Approach},
  author    = {Nguyen, Tuan and Honda, Hirotada and Sano, Takashi and Nguyen, Vinh and Nakamura, Shugo and Nguyen, Tan M.},
  booktitle = {Proceedings of The 27th International Conference on Artificial Intelligence and Statistics},
  series    = {Proceedings of Machine Learning Research},
  volume    = {238},
  pages     = {2710--2718},
  publisher = {PMLR},
  year      = {2024}
}

@inproceedings{song2025kuramoto_diffusion,
  title         = {Kuramoto Orientation Diffusion Models},
  author        = {Song, Yue and Keller, T. Anderson and Brodjian, Sevan and Miyato, Takeru and Yue, Yisong and Perona, Pietro and Welling, Max},
  booktitle     = {Advances in Neural Information Processing Systems},
  year          = {2025},
  eprint        = {2509.15328},
  archivePrefix = {arXiv},
  doi           = {10.48550/arXiv.2509.15328}
}

@article{breakspear2010generative,
  title   = {Generative Models of Cortical Oscillations: Neurobiological Implications of the Kuramoto Model},
  author  = {Breakspear, Michael and Heitmann, Stewart and Daffertshofer, Andreas},
  journal = {Frontiers in Human Neuroscience},
  volume  = {4},
  pages   = {190},
  year    = {2010},
  doi     = {10.3389/fnhum.2010.00190}
}

@article{ricci2021kuranet,
  title         = {KuraNet: Systems of Coupled Oscillators that Learn to Synchronize},
  author        = {Ricci, Matthew and Jung, Minju and Zhang, Yuwei and Chalvidal, Mathieu and Soni, Aneri and Serre, Thomas},
  journal       = {arXiv preprint arXiv:2105.02838},
  year          = {2021},
  eprint        = {2105.02838},
  archivePrefix = {arXiv},
  doi           = {10.48550/arXiv.2105.02838}
}

@article{xiao2026kope,
  title         = {Kuramoto Oscillatory Phase Encoding: Neuro-inspired Synchronization for Improved Learning Efficiency},
  author        = {Xiao, Mingqing and Wang, Yansen and Han, Dongqi and Shan, Caihua and Li, Dongsheng},
  journal       = {arXiv preprint arXiv:2604.07904},
  year          = {2026},
  eprint        = {2604.07904},
  archivePrefix = {arXiv},
  doi           = {10.48550/arXiv.2604.07904}
}

@article{ha2016emergent,
  title     = {Emergent dynamics of Winfree oscillators on locally coupled networks},
  author    = {Ha, Seung-Yeal and Ko, Dongnam and Park, Jinyeong and Ryoo, Sang Woo},
  journal   = {Journal of Differential Equations},
  volume    = {260},
  number    = {5},
  pages     = {4203--4236},
  year      = {2016},
  publisher = {Elsevier},
  doi       = {10.1016/j.jde.2015.11.008}
}

@article{manoranjani2023generalization,
  title   = {Generalization of the Kuramoto model to the Winfree model by a symmetry breaking coupling},
  author  = {Manoranjani, M. and Gupta, Shamik and Senthilkumar, D. V. and Chandrasekar, V. K.},
  journal = {The European Physical Journal Plus},
  volume  = {138},
  number  = {2},
  pages   = {144},
  year    = {2023},
  doi     = {10.1140/epjp/s13360-023-03760-5}
}

@article{hasson2008hierarchy,
  title   = {A hierarchy of temporal receptive windows in human cortex},
  author  = {Hasson, Uri and Yang, Eunice and Vallines, Ignacio and Heeger, David J. and Rubin, Nava},
  journal = {Journal of Neuroscience},
  volume  = {28},
  number  = {10},
  pages   = {2539--2550},
  year    = {2008},
  doi     = {10.1523/JNEUROSCI.5487-07.2008}
}

@article{honey2012slow,
  title   = {Slow cortical dynamics and the accumulation of information over long timescales},
  author  = {Honey, Christopher J. and Thesen, Thomas and Donner, Tobias H. and Silbert, Lauren J. and Carlson, Chad E. and Devinsky, Orrin and Doyle, Werner K. and Rubin, Nava and Heeger, David J. and Hasson, Uri},
  journal = {Neuron},
  volume  = {76},
  number  = {2},
  pages   = {423--434},
  year    = {2012},
  doi     = {10.1016/j.neuron.2012.08.011}
}

@article{golesorkhi2021temporal,
  title   = {Temporal hierarchy of intrinsic neural timescales converges with spatial core-periphery organization},
  author  = {Golesorkhi, Mehrshad and Gomez-Pilar, Javier and Tumati, Shankar and Fraser, Maia and Northoff, Georg},
  journal = {Communications Biology},
  volume  = {4},
  number  = {1},
  pages   = {277},
  year    = {2021},
  doi     = {10.1038/s42003-021-01785-z}
}

@article{wang2025hrm,
  title         = {Hierarchical Reasoning Model},
  author        = {Wang, Guan and Li, Jin and Sun, Yuhao and Chen, Xing and Liu, Changling and Wu, Yue and Lu, Meng and Song, Sen and Abbasi Yadkori, Yasin},
  journal       = {arXiv preprint arXiv:2506.21734},
  year          = {2025},
  eprint        = {2506.21734},
  archivePrefix = {arXiv},
  primaryClass  = {cs.AI}
}

@techreport{krizhevsky2009learning,
  title       = {Learning Multiple Layers of Features from Tiny Images},
  author      = {Krizhevsky, Alex},
  institution = {University of Toronto},
  year        = {2009}
}

@inproceedings{deng2009imagenet,
  title     = {ImageNet: A Large-Scale Hierarchical Image Database},
  author    = {Deng, Jia and Dong, Wei and Socher, Richard and Li, Li-Jia and Li, Kai and Fei-Fei, Li},
  booktitle = {Proceedings of the IEEE Conference on Computer Vision and Pattern Recognition},
  pages     = {248--255},
  year      = {2009},
  doi       = {10.1109/CVPR.2009.5206848}
}

@inproceedings{he2016deep,
  title     = {Deep Residual Learning for Image Recognition},
  author    = {He, Kaiming and Zhang, Xiangyu and Ren, Shaoqing and Sun, Jian},
  booktitle = {Proceedings of the IEEE Conference on Computer Vision and Pattern Recognition},
  pages     = {770--778},
  year      = {2016},
  doi       = {10.1109/CVPR.2016.90}
}

@inproceedings{dosovitskiy2021image,
  title     = {An Image is Worth 16x16 Words: Transformers for Image Recognition at Scale},
  author    = {Dosovitskiy, Alexey and Beyer, Lucas and Kolesnikov, Alexander and Weissenborn, Dirk and Zhai, Xiaohua and Unterthiner, Thomas and Dehghani, Mostafa and Minderer, Matthias and Heigold, Georg and Gelly, Sylvain and Uszkoreit, Jakob and Houlsby, Neil},
  booktitle = {International Conference on Learning Representations},
  year      = {2021}
}

@inproceedings{touvron2021training,
  title     = {Training Data-Efficient Image Transformers and Distillation through Attention},
  author    = {Touvron, Hugo and Cord, Matthieu and Douze, Matthijs and Massa, Francisco and Sablayrolles, Alexandre and J{\'e}gou, Herv{\'e}},
  booktitle = {International Conference on Machine Learning},
  pages     = {10347--10357},
  year      = {2021}
}

@article{deepseekai2025deepseekr1,
  title   = {DeepSeek-R1 incentivizes reasoning in LLMs through reinforcement learning},
  author  = {Guo, Daya and Yang, Dejian and Zhang, Haowei and Song, Junxiao and Wang, Peiyi and Zhu, Qihao and Xu, Runxin and Zhang, Ruoyu and Ma, Shirong and Bi, Xiao and others},
  journal = {Nature},
  volume  = {645},
  pages   = {633--638},
  year    = {2025},
  doi     = {10.1038/s41586-025-09422-z}
}

@article{jolicoeurmartineau2025trm,
  title         = {Less is More: Recursive Reasoning with Tiny Networks},
  author        = {Jolicoeur-Martineau, Alexia},
  journal       = {arXiv preprint arXiv:2510.04871},
  year          = {2025},
  eprint        = {2510.04871},
  archivePrefix = {arXiv},
  primaryClass  = {cs.LG},
  doi           = {10.48550/arXiv.2510.04871}
}

@inproceedings{lehnert2024beyond,
  title     = {Beyond A*: Better Planning with Transformers via Search Dynamics Bootstrapping},
  author    = {Lehnert, Lucas and Sukhbaatar, Sainbayar and Su, DiJia and Zheng, Qinqing and McVay, Paul and Rabbat, Michael and Tian, Yuandong},
  booktitle = {First Conference on Language Modeling},
  year      = {2024}
}

@inproceedings{wang2019satnet,
  title     = {{SATNet}: Bridging Deep Learning and Logical Reasoning Using a Differentiable Satisfiability Solver},
  author    = {Wang, Po-Wei and Donti, Priya and Wilder, Bryan and Kolter, Zico},
  booktitle = {Proceedings of the 36th International Conference on Machine Learning},
  pages     = {6545--6554},
  year      = {2019}
}

@inproceedings{palm2018recurrent,
  title     = {Recurrent Relational Networks},
  author    = {Palm, Rasmus Berg and Paquet, Ulrich and Winther, Ole},
  booktitle = {Advances in Neural Information Processing Systems},
  volume    = {31},
  year      = {2018}
}

@inproceedings{yang2023recurrent,
  title     = {Learning to Solve Constraint Satisfaction Problems with Recurrent Transformer},
  author    = {Yang, Zhun and Ishay, Adam and Lee, Joohyung},
  booktitle = {International Conference on Learning Representations},
  year      = {2023}
}

@book{buzsaki2006rhythms,
  title     = {Rhythms of the Brain},
  author    = {Buzs{\'a}ki, Gy{\"o}rgy},
  year      = {2006},
  publisher = {Oxford University Press},
  address   = {New York}
}

@article{fries2005mechanism,
  title   = {A Mechanism for Cognitive Dynamics: Neuronal Communication Through Neuronal Coherence},
  author  = {Fries, Pascal},
  journal = {Trends in Cognitive Sciences},
  volume  = {9},
  number  = {10},
  pages   = {474--480},
  year    = {2005},
  doi     = {10.1016/j.tics.2005.08.011}
}

@article{muller2018cortical,
  title   = {Cortical Travelling Waves: Mechanisms and Computational Principles},
  author  = {Muller, Lyle and Chavane, Fr{\'e}d{\'e}ric and Reynolds, John and Sejnowski, Terrence J.},
  journal = {Nature Reviews Neuroscience},
  volume  = {19},
  number  = {5},
  pages   = {255--268},
  year    = {2018},
  doi     = {10.1038/nrn.2018.20}
}

@article{effenberger2025functional,
  title   = {The Functional Role of Oscillatory Dynamics in Neocortical Circuits: A Computational Perspective},
  author  = {Effenberger, Felix and Carvalho, Pedro and Dubinin, Igor and Singer, Wolf},
  journal = {Proceedings of the National Academy of Sciences},
  volume  = {122},
  number  = {4},
  pages   = {e2412830122},
  year    = {2025},
  doi     = {10.1073/pnas.2412830122}
}

@inproceedings{neil2016phased,
  title     = {Phased {LSTM}: Accelerating Recurrent Network Training for Long or Event-based Sequences},
  author    = {Neil, Daniel and Pfeiffer, Michael and Liu, Shih-Chii},
  booktitle = {Advances in Neural Information Processing Systems},
  volume    = {29},
  year      = {2016}
}

@inproceedings{rusch2021cornn,
  title     = {Coupled Oscillatory Recurrent Neural Network ({coRNN}): An Accurate and (Gradient) Stable Architecture for Learning Long Time Dependencies},
  author    = {Rusch, T. Konstantin and Mishra, Siddhartha},
  booktitle = {International Conference on Learning Representations},
  year      = {2021}
}

@inproceedings{rusch2022graph,
  title     = {Graph-Coupled Oscillator Networks},
  author    = {Rusch, T. Konstantin and Chamberlain, Benjamin P. and Rowbottom, James and Mishra, Siddhartha and Bronstein, Michael M.},
  booktitle = {Proceedings of the 39th International Conference on Machine Learning},
  series    = {Proceedings of Machine Learning Research},
  volume    = {162},
  pages     = {18888--18909},
  publisher = {PMLR},
  year      = {2022}
}

@inproceedings{keller2023neural,
  title     = {Neural Wave Machines: Learning Spatiotemporally Structured Representations with Locally Coupled Oscillatory Recurrent Neural Networks},
  author    = {Keller, T. Anderson and Welling, Max},
  booktitle = {Proceedings of the 40th International Conference on Machine Learning},
  series    = {Proceedings of Machine Learning Research},
  volume    = {202},
  pages     = {16168--16189},
  publisher = {PMLR},
  year      = {2023}
}

@inproceedings{keller2024traveling,
  title     = {Traveling Waves Encode the Recent Past and Enhance Sequence Learning},
  author    = {Keller, T. Anderson and Muller, Lyle and Sejnowski, Terrence J. and Welling, Max},
  booktitle = {International Conference on Learning Representations},
  year      = {2024}
}

@article{duecker2024oscillations,
  title   = {Oscillations in an Artificial Neural Network Convert Competing Inputs into a Temporal Code},
  author  = {Duecker, Katharina and Idiart, Marco and van Gerven, Marcel and Jensen, Ole},
  journal = {PLOS Computational Biology},
  volume  = {20},
  number  = {9},
  pages   = {e1012429},
  year    = {2024},
  doi     = {10.1371/journal.pcbi.1012429}
}

@book{yue2025structured,
  title={Structured representation learning: from homomorphisms and disentanglement to equivariance and topography},
  author={Song, Yue and Keller, Thomas Anderson and Sebe, Nicu and Welling, Max},
  year={2025},
  publisher={Springer Nature}
}

@inproceedings{song2023flow,
  title={Flow Factorized Representation Learning},
  author={Song, Yue and Keller, Andy and Sebe, Nicu and Welling, Max},
  booktitle={Advances in Neural Information Processing Systems},
  volume={36},
  year={2023}
}

@inproceedings{song2023latent,
  title={Latent Traversals in Generative Models as Potential Flows},
  author={Song, Yue and Keller, Andy and Sebe, Nicu and Welling, Max},
  booktitle={International Conference on Machine Learning},
  year={2023},
  organization={PMLR}
}

@article{tian2025dassdp,
  title   = {Synchrony-Gated Plasticity with Dopamine Modulation for Spiking Neural Networks},
  author  = {Tian, Yuchen and Tensingh, Samuel and Eshraghian, Jason K. and Truong, Nhan Duy and Kavehei, Omid},
  journal = {Transactions on Machine Learning Research},
  year    = {2025}
}

@article{darlow2025ctm,
  title   = {Continuous Thought Machines},
  author  = {Darlow, Luke and Regan, Ciaran and Risi, Sebastian and Seely, Jeffrey and Jones, Llion},
  journal = {arXiv preprint arXiv:2505.05522},
  year    = {2025}
}

\appendix

\clearpage
\appendix

\section*{Appendix Contents}
\startcontents[appendix]
\printcontents[appendix]{l}{1}{\setcounter{tocdepth}{3}}

\clearpage
\section{Related Work}
\noindent\textbf{Oscillatory neural computation.}
Oscillatory activity is widely regarded as a fundamental organizing principle in biological neural systems, supporting temporal coordination, communication, and structured information processing~\citep{buzsaki2006rhythms,fries2005mechanism,muller2018cortical}. This perspective has motivated the use of oscillatory dynamics as inductive biases in artificial neural networks, where oscillations provide a temporal scaffold for computation. Early work introduced rhythmic gating mechanisms for sparse and event-based processing~\citep{neil2016phased}, while more recent approaches have explored dynamical systems-based architectures, including stable recurrent networks and graph-based models, where oscillatory or wave-like dynamics enable long-range interactions and memory~\citep{rusch2021cornn,rusch2022graph,keller2023neural,song2023latent,song2023flow,keller2024traveling}. More broadly, oscillatory mechanisms have been shown to improve learning dynamics, robustness, and representational organization by structuring computation through phase and temporal evolution~\citep{duecker2024oscillations,effenberger2025functional,yue2025structured}.

\noindent\textbf{Synchronization-inspired models in machine learning.}
While oscillations structure computation in time, synchronization governs how multiple oscillators interact through collective phase alignment. Building on this principle, a growing body of work has incorporated synchronization into neural architectures. 
Prior approaches have leveraged phase or complex-valued representations to organize object-centric features and latent structure~\citep{lowe2022cae,stanic2023ctcae,lowe2023rotating,gopalakrishnan2024syncx}, introduced synchronization mechanisms into modern architectures through bounded-confidence interactions~\citep{liu2026krause}, and explored neural synchronization as a latent representation in recurrent dynamical architectures such as the Continuous Thought Machine~\citep{darlow2025ctm}. Synchrony-based signals have also been used as training-time plasticity mechanisms in spiking neural networks, for example through dopamine-modulated spike-synchrony-dependent plasticity~\citep{tian2025dassdp}. Closely related to our work are oscillator-based neural systems such as \emph{AKOrN}~\citep{miyato2025akorn}.
At a more fundamental level, these approaches are closely related to classical coupled-oscillator systems, particularly the Kuramoto and Winfree models~\citep{kuramoto1984chemical,winfree1967biological}. Kuramoto-type dynamics based on pairwise phase interactions have been widely explored in cortical oscillation modeling, learnable oscillator systems, graph neural networks, orientation diffusion models, and phase-based neural architectures~\citep{breakspear2010generative,ricci2021kuranet,nguyen2024kuramatognn,song2025kuramoto_diffusion,xiao2026kope}. In contrast, Winfree-type formulations introduce separable sensitivity--influence interactions, enabling more flexible and expressive collective dynamics~\citep{ha2016emergent,manoranjani2023generalization}. \emph{WONN} builds on this formulation by using generalized Winfree dynamics with flexible parameterizations and grouped interactions, enabling hierarchical information flow through synchronization dynamics on a toroidal phase space.

\section{Discussions}

\subsection{Limitation and Future Work}

Our results suggest that oscillatory synchronization can serve not only as a biological or physical metaphor, but also as a practical computational principle for modern neural networks. At the same time, \emph{WONN} represents only an initial step toward synchronization-driven large-scale neural computation. While the recurrent Winfree dynamics provide strong parameter efficiency and expressive iterative refinement, they also introduce additional computational cost compared with conventional feed-forward architectures such as ResNets and ViTs, since each layer involves multiple recurrent synchronization steps. Furthermore, although \emph{WONN} demonstrates competitive scalability up to ImageNet-1K, exploring larger-scale models and even language models remains an important direction for future work. Future research may further investigate more efficient synchronization mechanisms, adaptive or continuous-time dynamics, larger-scale oscillatory architectures, and deeper theoretical understanding of the stability, geometry, and representation structure induced by Winfree-type neural dynamics.

\subsection{Difference against \emph{AKOrN}}
\emph{WONN} is closely related to \emph{AKOrN}~\cite{miyato2025akorn} in that both methods introduce oscillatory dynamics into neural architectures. 
However, the two models differ in several essential aspects.

\noindent\textbf{Practical scalability.} In Table~\ref{tab:imagenet1k_impl}, we additionally compare the practical training cost of \emph{WONN} and \emph{AKOrN} on ImageNet-1K. Despite using 8 poweful H200 GPUs, \emph{AKOrN} requires nearly 1,000 GPU hours to train its standard-size model for only 200 epochs, whereas \emph{WONN} is trained for 300 epochs using only 4 H100 GPUs with substantially lower total GPU hours. We also observe a large memory gap under the same per-GPU batch size: \emph{AKOrN} requires more than 110GB aggregate GPU memory, compared with approximately 35GB for \emph{WONN}. These results suggest that \emph{WONN} is considerably more favorable for large-scale training under realistic compute and memory constraints.

\noindent\textbf{Different state geometries.}
\emph{AKOrN} represents each oscillatory neuron by a high-dimensional unit vector, so its oscillator state is embedded in a Euclidean ambient space. In contrast, \emph{WONN} represents the oscillator state directly by a phase angle. Thus, each scalar oscillator evolves on the circle $S^1$, and a collection of $d$ oscillators naturally lives on the torus $(S^1)^d$. This makes the periodic geometry of the representation explicit rather than treating oscillation through a vector-valued Euclidean embedding.

\noindent\textbf{Separable interaction mechanisms.}
\emph{AKOrN} is motivated by Kuramoto-type synchronization (Eq.~\ref{eq:kuramoto_model}), where interactions are governed primarily by pairwise phase differences. Such dynamics are invariant under global phase shifts, since replacing every phase by $\theta_i+\alpha$ leaves the differences $\theta_j-\theta_i$ unchanged. While this symmetry simplifies the interaction structure, it may also restrict the class of admissible dynamics. In contrast, \emph{WONN} builds upon the more general Winfree formulation (Eq.~\ref{eq:winfree}), where oscillator updates are decomposed into a sensitivity function of the receiving oscillator and an influence function of neighboring oscillators. This formulation recovers Kuramoto-style phase-difference coupling as a special case, while also permitting symmetry-breaking interactions (Eq.~\ref{eq:generalized_winfree}) that depend on absolute phase. As a result, \emph{WONN} defines a more flexible and expressive class of oscillatory interactions.

\noindent\textbf{Grouped hierarchical interactions.}
\emph{WONN} extends the classical Winfree formulation into a grouped neural architecture. Instead of applying only pointwise oscillator interactions, \emph{WONN} partitions oscillators into groups and computes group-level influence signals. 
This grouped formulation induces a hierarchical interaction structure: oscillators may synchronize locally within a group, while group-level influence signals coordinate information across larger spatial or feature regions. The group size therefore controls the interaction scale, interpolating between pointwise dynamics and more structured collective dynamics.

\noindent\textbf{Flexible interaction parameterization.} \emph{WONN} provides a flexible parameterization of the Winfree interaction functions. The sensitivity and influence functions can either be instantiated as fixed trigonometric mappings, such as $S(\theta)=\cos\theta$ and $I(\theta)=\sin\theta$, or implemented as learnable neural networks. The former preserves a strong oscillator-inspired inductive bias, while the latter increases representational flexibility and expressiveness. This allows the interaction dynamics of \emph{WONN} to adapt naturally to different tasks, datasets, and computational regimes.

\section{Additional Implementation Details and Extended Experiment Results}
\label{sec:implementation_results_app}

\subsection{Implementation Details} 
\label{sec:Implementation Train Details}

\makeatletter
\renewcommand\subsubsection{\@startsection{subsubsection}{3}{\z@}%
  {-3.25ex\@plus -1ex \@minus -.2ex}%
  {1.5ex \@plus .2ex}%
  {\normalfont\normalsize\scshape}}
\makeatother

\subsubsection{{Image Recognition}}

For Image Recognition, we train all the models under comparable training protocols. The detailed training recipes are reported in Table~\ref{tab:cifar_impl} for \textbf{CIFAR-10/100}, Table~\ref{tab:imagenet100_impl} for \textbf{ImageNet-100}, and Table ~\ref{tab:imagenet1k_impl} for \textbf{ImageNet-1K}.


\FloatBarrier
\noindent
\begin{minipage}{\columnwidth}
    \centering
    \captionsetup{hypcap=false}
    \captionof{table}{Implementation details on \textbf{CIFAR-10/100}.}
    \label{tab:cifar_impl}
    \small
    \setlength{\tabcolsep}{3.5pt}
    \begin{tabular}{l|ccccc}
        \toprule
        \textbf{Hyperparameters}
        & \multicolumn{5}{c}{\textbf{Models}} \\
        \cmidrule(lr){2-6}
        & \textbf{\emph{WONN}}
        & \textbf{\emph{AKOrN}$^{\mathrm{attn}}$}
        & \textbf{ResNet-18/50}
        & \textbf{ViT-T/S}
        & \textbf{ViT-B} \\
        \midrule
        Batch Size        & 64 & 128 & 128 & 256 & 256 \\
        Optimizer         & AdamW & Adam & AdamW & AdamW & AdamW \\
        Weight Decay      & 0.0001 & --  & 0.0001 & 0.05 & 0.05 \\
        LR                & $5 \cdot 10^{-4}$ & $5 \cdot 10^{-4}$ & $5 \cdot 10^{-4}$ & $3 \cdot 10^{-4}$ & $1 \cdot 10^{-4}$ \\
        LR Decay          & \xmark & \xmark & \xmark & cosine & cosine \\
        Min LR            & \xmark & \xmark & \xmark & 0 & 0 \\
        Warmup Epochs     & \xmark & \xmark & \xmark & 10 & 10 \\
        Epochs            & 200 & 200 & 200 & 300 & 300 \\
        Dynamics Layers (L) & 6 & 3 & -- & -- & -- \\
        Dynamics Time Step (T) & 3 & 3 & -- & -- & -- \\
        Group Size (N)   & 2 & -- & -- & -- & -- \\
        Oscillator Dimension (n)   & -- & 4 & -- & -- & -- \\
        GPUs              & 1$\times$H100 & 1$\times$H100 & 1$\times$H100 & 1$\times$H100 & 1$\times$H100 \\
        Dropout           & \xmark & \xmark & \xmark & \xmark & \xmark \\
        Label Smoothing   & \xmark & \xmark & \xmark & 0.1 & 0.1 \\
        Mixup Alpha       & \xmark & \xmark & \xmark & 0.8 & 0.8 \\
        Cutmix Alpha      & \xmark & \xmark & \xmark & 1.0 & 1.0 \\
        Mixup Prob.       & \xmark & \xmark & \xmark & 0.2 & 0.2 \\
        Cutmix Prob.       & \xmark & \xmark & \xmark & 0.8 & 0.8 \\
        \bottomrule
    \end{tabular}
\end{minipage}
\FloatBarrier

\FloatBarrier
\noindent
\begin{minipage}{\columnwidth}
    \centering
    \captionsetup{hypcap=false}
    \captionof{table}{Implementation details on \textbf{ImageNet-100}.}
    \label{tab:imagenet100_impl}
    \small
    \setlength{\tabcolsep}{3.0pt}
    \begin{tabular}{l|cccccc}
        \toprule
        \textbf{Hyperparameters}
        & \multicolumn{6}{c}{\textbf{Models}} \\
        \cmidrule(lr){2-7}
        & \textbf{\emph{WONN}}
        & \textbf{\emph{AKOrN}$^{\mathrm{attn}}$}
        & \textbf{ResNet-18}
        & \textbf{ResNet-50}
        & \textbf{ViT-S-16}
        & \textbf{ViT-B-16} \\
        \midrule
        Batch Size        & 64 & 16 & 128 & 128 & 256 & 256 \\
        Optimizer         & AdamW & Adam & AdamW & AdamW & AdamW & AdamW \\
        Weight Decay      & 0.0001 & -- & 0.0001 & 0.0001 & 0.05 & 0.05 \\
        LR                & $5 \cdot 10^{-4}$ & $5 \cdot 10^{-4}$ & $5 \cdot 10^{-4}$ & $5 \cdot 10^{-4}$ & $3 \cdot 10^{-4}$ & $2/3 \cdot 10^{-4}$ \\
        LR Decay          & \xmark & \xmark & \xmark & \xmark & cosine & cosine \\
        Min LR            & \xmark & \xmark & \xmark & \xmark & 0 & 0 \\
        Warmup Epochs     & \xmark & \xmark & \xmark & \xmark & 10 & 10 \\
        Epochs            & 200 & 200 & 200 & 200 & 300 & 300 \\
        Dynamics Layers ($L$) & 6 & 3 & -- & -- & -- & -- \\
        Dynamics Time Step ($T$) & 3 & 3 & -- & -- & -- & -- \\
        Group Size ($N$)   & 2 & -- & -- & -- & -- & -- \\
        Oscillator Dimension ($n$)   & -- & 4 & -- & -- & -- & -- \\
        Input Patch Size  & 4 & 4 & -- & -- & 16 & 16 \\
        GPUs              & 1$\times$H100 & 1$\times$H100 & 1$\times$H100 & 1$\times$H100 & 1$\times$H100 & 1$\times$H100 \\
        Dropout           & \xmark & \xmark & \xmark & \xmark & \xmark & \xmark \\
        Label Smoothing   & \xmark & \xmark & \xmark & \xmark & 0.1 & 0.1 \\
        Mixup Alpha       & \xmark & \xmark & \xmark & \xmark & 0.8 & 0.8 \\
        Cutmix Alpha      & \xmark & \xmark & \xmark & \xmark & 1.0 & 1.0 \\
        Mixup Prob.       & \xmark & \xmark & \xmark & \xmark & 0.2 & 0.2 \\
        Cutmix Prob.      & \xmark & \xmark & \xmark & \xmark & 0.8 & 0.8 \\
        \bottomrule
    \end{tabular}
\end{minipage}
\FloatBarrier

\FloatBarrier
\noindent
\begin{minipage}{\columnwidth}
    \centering
    \captionsetup{hypcap=false}
    \captionof{table}{Implementation details on \textbf{ImageNet-1K}.}
    \label{tab:imagenet1k_impl}
    \small
    \setlength{\tabcolsep}{3.0pt}
    \begin{tabular}{l|cccccc}
        \toprule
        \textbf{Hyperparameters}
        & \multicolumn{6}{c}{\textbf{Models}} \\
        \cmidrule(lr){2-7}
        & \textbf{\emph{WONN}}
        & \textbf{\emph{AKOrN}$^{\mathrm{attn}}$}
        & \textbf{ResNet-18}
        & \textbf{ResNet-50}
        & \textbf{ViT-S-16}
        & \textbf{ViT-B-16} \\
        \midrule
        Batch Size        & 512 & 512 & 128 & 128 & 1024 & 1024 \\
        Optimizer         & AdamW & AdamW & AdamW & AdamW & AdamW & AdamW \\
        Weight Decay      & 0.001 & 0.001 & 0.001 & 0.001 & 0.05 & 0.05 \\
        LR                & $7.5 \cdot 10^{-4}$ & $5 \cdot 10^{-4}$ & $5 \cdot 10^{-4}$ & $5 \cdot 10^{-4}$ & $3 \cdot 10^{-4}$ & $2/3 \cdot 10^{-4}$ \\
        LR Decay          & cosine & cosine & cosine & cosine & cosine & cosine \\
        Min LR            & $1 \cdot 10^{-6}$ & $1 \cdot 10^{-6}$ & $1 \cdot 10^{-6}$ & $1 \cdot 10^{-6}$ & 0 & 0 \\
        Warmup Epochs     & 10 & 10 & 10 & 10 & 10 & 10 \\
        Epochs            & 300 & 200 & 200 & 200 & 300 & 300 \\
        Dynamics Layers ($L$) & 6 & 3 & -- & -- & -- & -- \\
        Dynamics Time Step ($T$) & 3 & 3 & -- & -- & -- & -- \\
        Group Size ($N$)   & 2 & -- & -- & -- & -- & -- \\
        Oscillator Dimension ($n$) & -- & 4 & -- & -- & -- & -- \\
        Input Patch Size  & 7 & 7 & -- & -- & 16 & 16 \\
        GPUs              & $4\times$H100 & $8\times$H200 & $1\times$H100 & $1\times$H100 & $4\times$H100 & $4\times$H100 \\
        GPU hours              & 750 & 950 & -- & -- & -- & -- \\
        Dropout           & \xmark & \xmark & \xmark & \xmark & \xmark & \xmark \\
        Mixup Alpha       & -- & -- & -- & -- & 0.8 & 0.8 \\
        Cutmix Alpha      & 1.0 & 1.0 & 1.0 & 1.0 & 1.0 & 1.0 \\
        Label Smoothing   & 0.1 & 0.1 & 0.1 & 0.1 & 0.1 & 0.1 \\
        Mixup Prob.       & 0.0 & 0.0 & 0.0 & 0.0 & 0.2 & 0.2 \\
        Cutmix Prob.      & 0.8 & 0.8 & 0.8 & 0.8 & 0.8 & 0.8 \\
        \bottomrule
    \end{tabular}
\end{minipage}
\FloatBarrier

\subsubsection{{Solving Maze-hard}}
For the Maze-hard task, we train the two models \emph{WONN} and \emph{AKOrN} under the same training protocol, with implementation details summarized in Table~\ref{tab:maze_impl}. Results for the remaining baselines, including large language models and recurrent models, are taken from~\citep{wang2025hrm,jolicoeurmartineau2025trm}.

\FloatBarrier
\noindent
\begin{minipage}{\columnwidth}
    \centering
    \captionsetup{hypcap=false}
    \captionof{table}{Implementation details on \textbf{Maze-hard}.}
    \label{tab:maze_impl}
    \small
    \setlength{\tabcolsep}{6pt}
    \begin{tabular}{l|cc}
        \toprule
        \textbf{Hyperparameters}
        & \multicolumn{2}{c}{\textbf{Models}} \\
        \cmidrule(lr){2-3}
        & \textbf{\emph{WONN}}
        & \textbf{\emph{AKOrN}$^{\mathrm{attn}}$} \\
        \midrule
        Batch Size              & 64 & 64 \\
        Channel                 &128 & 256 \\
        Optimizer               & Adam & Adam \\
        Weight Decay            & -- & -- \\
        LR                      & $1 \cdot 10^{-3}$ & $1 \cdot 10^{-3}$ \\
        LR Decay                & -- & -- \\
        Min LR                  & -- & -- \\
        Warmup Epochs           & -- & -- \\
        Epochs                  & 9000 & 9000 \\
        Dynamics Layers ($L$)   & 1 & 1 \\
        Dynamics Time Step ($T$)& 24 & 24 \\
        Group Size ($N$)        & 1 & -- \\
        GPUs                    & $1\times$H100 & $1\times$H100 \\
        \bottomrule
    \end{tabular}
\end{minipage}
\FloatBarrier

\subsubsection{{Solving Sudoku}}

For the Sudoku task, we train \emph{WONN} following the same training protocol as \emph{AKOrN}. The detailed configuration is summarized in Table~\ref{tab:sudoku_impl}. The \emph{HRM} results are obtained from our reimplementation, while results for the remaining baselines are taken from prior work~\citep{wang2019satnet,palm2018recurrent,miyato2025akorn}.

\FloatBarrier
\noindent
\begin{minipage}{\columnwidth}
    \centering
    \captionsetup{hypcap=false}
    \captionsetup{hypcap=false}
    \captionof{table}{Implementation details on \textbf{Sudoku}.}
    \label{tab:sudoku_impl}
    \small
    \setlength{\tabcolsep}{5pt}
    \begin{tabular}{l|ccc}
        \toprule
        \textbf{Hyperparameters}
        & \multicolumn{3}{c}{\textbf{Models}} \\
        \cmidrule(lr){2-4}
        & \textbf{\emph{WONN}}
        & \textbf{\emph{AKOrN}$^{\mathrm{attn}}$}
        & \textbf{\emph{HRM}} \\
        \midrule
        Batch Size              & 100 & 100 & 384 \\
        Channel                 & 256 & 512 & 512 \\
        Optimizer               & Adam & Adam & Adam-atan2 \\
        Weight Decay            & -- & -- & 1.0 \\
        LR                      & $1 \cdot 10^{-3}$ & $1 \cdot 10^{-3}$ & $1 \cdot 10^{-4}$ \\
        LR Decay                & -- & -- & -- \\
        Min LR                  & -- & -- & -- \\
        Warmup Epochs           & -- & -- & 2000 steps \\
        Epochs                  & 100 & 100 & 2000 \\
        Dynamics Layers ($L$)   & 1 & 1 & $H/L=4/4$ \\
        Dynamics Time Step ($T$)& 16 & 16 & -- \\
        Group Size ($N$)        & 1 & -- & -- \\
        GPUs                    & $1\times$H100 & $1\times$H100 & $1\times$H100 \\
        \bottomrule
    \end{tabular}
\end{minipage}
\FloatBarrier

\subsection{Additional Results and Ablations on Image Recognition}
\label{sec:additional_image_recognition_results_and_ablations}

Here we provide additional experimental results for image recognition benchmarks, including supplementary evaluations on ImageNet-100 (Table~\ref{tab:imagenet100_addition}) and CIFAR-100 (Table~\ref{tab:cifar100_addition}). We further conduct several ablation studies on CIFAR-100 to analyze the contribution of key architectural and dynamical components of \emph{WONN}. Throughout this section, $L$ denotes the number of Winfree dynamics layers, $T$ denotes the number of recurrent dynamics steps per layer, and $N$ denotes the group size. Tables~\ref{tab:imagenet100_addition} and~\ref{tab:cifar100_addition} summarize the ablation results on ImageNet-100 and CIFAR-100, respectively.

\FloatBarrier
\noindent
\begin{minipage}{\columnwidth}
    \centering
    \captionsetup{hypcap=false}
    \captionof{table}{Additional image classification results on \textbf{ImageNet-100}.}
    \label{tab:imagenet100_addition}
    \small
    \setlength{\tabcolsep}{4pt}
    \begin{tabular}{l|cc}
        \toprule
        \textbf{Models}
        & \textbf{Accuracy (\%)} 
        & \textbf{\# Parameters} \\
        \midrule
        \emph{\textbf{AKOrN}}$^{\mathrm{attn}}$ (input patch size = 7)  
        & 75.12 & 4.63M \\
        \emph{\textbf{AKOrN}}$^{\mathrm{attn}}$ (input patch size = 4)  
        & 80.08 & 4.62M \\
        \midrule
        \multicolumn{3}{l}{\textit{$S(\theta_i)$ and $I(\theta_i)$ as MLPs, input patch size = 8}}\\
        \emph{\textbf{WONN}} ($\mathrm{Ch}=256 \to 256$) 
        & 82.08 & 12.09M \\
        \midrule
        \multicolumn{3}{l}{\textit{$S(\theta_i)$ and $I(\theta_i)$ as MLPs, input patch size = 7}}\\
        \emph{\textbf{WONN}} ($\mathrm{Ch}=128 \to 128$) 
        & 78.78 & 3.11M \\
        \emph{\textbf{WONN}} ($\mathrm{Ch}=64 \to 256$) 
        & 79.12 & 7.57M \\
        \emph{\textbf{WONN}} ($\mathrm{Ch}=256 \to 256$) 
        & 80.08 & 12.08M \\
        \midrule 
        \multicolumn{3}{l}{\textit{$S(\theta_i)$ and $I(\theta_i)$ as MLPs, input patch size = 4}}\\
        \emph{\textbf{WONN}} ($\mathrm{Ch}=128 \to 128$)
        & 81.50 & 3.09M \\
        \emph{\textbf{WONN}} ($\mathrm{Ch}=64 \to 256$) 
        & 82.04 & 7.56M \\
        \emph{\textbf{WONN}} ($\mathrm{Ch}=256 \to 256$) 
        & 82.88 & 12.05M \\
        \midrule 
        \multicolumn{3}{l}{\textit{$S(\theta_i)$ and $I(\theta_i)$ as trigonometric functions, input patch size = 7}}\\
        \emph{\textbf{WONN}} ($\mathrm{Ch}=128 \to 128$) 
        & 77.92 & 3.01M \\
        \emph{\textbf{WONN}} ($\mathrm{Ch}=64 \to 256$) 
        & 78.78 & 7.44M \\
        \emph{\textbf{WONN}} ($\mathrm{Ch}=256 \to 256$) 
        & 80.34 & 11.89M \\
        \midrule 
        \multicolumn{3}{l}{\textit{$S(\theta_i)$ and $I(\theta_i)$ as trigonometric functions, input patch size = 4}}\\
        \emph{\textbf{WONN}} ($\mathrm{Ch}=128 \to 128$) & 81.76 & 3.00M \\
        \emph{\textbf{WONN}} (($\mathrm{Ch}=64 \to 256$)  & 82.56 & 7.43M\\
        \emph{\textbf{WONN}} ($\mathrm{Ch}=256 \to 256$)  & 82.22 & 11.87M\\
        \bottomrule
    \end{tabular}
\end{minipage}
\FloatBarrier

\FloatBarrier
\noindent
\begin{minipage}{\columnwidth}
    \centering
    \captionsetup{hypcap=false}
    \captionof{table}{Additional image classification results on \textbf{CIFAR-100}.}
    \label{tab:cifar100_addition}
    \small
    \setlength{\tabcolsep}{5pt}
    \begin{tabular}{l|cc}
        \toprule
        \textbf{Models}
        & \textbf{Accuracy (\%)}
        & \textbf{$\#$ Parameters} \\
        \midrule
        \multicolumn{3}{l}{\textit{$S(\theta_i)\&I(\theta_i)$ as MLPs, $L=6$, $T=3$, $N=2$}}\\
        \emph{\textbf{WONN}} ($\mathrm{Ch}=128 \to 128$) 
        & 73.77{\tiny$\pm$0.40} & 3.09M \\
        \emph{\textbf{WONN}} ($\mathrm{Ch}=64 \to 256$) 
        & 75.12{\tiny$\pm$0.49} & 7.56M \\
        \emph{\textbf{WONN}} ($\mathrm{Ch}=256 \to 256$) 
        & 76.20{\tiny$\pm$0.45} & 12.04M \\
        \midrule
        \multicolumn{3}{l}{\textit{$S(\theta_i)\&I(\theta_i)$ as trigonometric functions, $L=6$, $T=3$, $N=2$}}\\
        \emph{\textbf{WONN}} ($\mathrm{Ch}=128 \to 128$) 
        & 74.48{\tiny$\pm$0.16} & 3.00M \\
        \emph{\textbf{WONN}} ($\mathrm{Ch}=64 \to 256$) 
        & 75.81{\tiny$\pm$0.33} & 7.43M \\
        \emph{\textbf{WONN}} ($\mathrm{Ch}=256 \to 256$) 
        & 76.17{\tiny$\pm$0.52} & 11.86M \\
        \midrule
        \multicolumn{3}{l}{\textit{$S(\theta_i)\&I(\theta_i)$ as MLPs, Ch=256, local (conv) and globle (attn) coupling}}\\
        \emph{\textbf{WONN}}$^{attn}$ (L1T3N2) 
        & 66.14 & 1.65M \\
        \emph{\textbf{WONN}}$^{attn}$ (L6T1N2) 
        & 74.53 & 12.04M \\
        \emph{\textbf{WONN}}$^{attn}$ (L6T3N1) 
        & 75.93 & 11.82M \\
        \emph{\textbf{WONN}}$^{attn}$ (L6T3N2)
        & 76.20 & 12.04M \\
        \emph{\textbf{WONN}}$^{conv}$ (L6T3N2) 
        & 75.63 & 18.72M \\
        \emph{\textbf{WONN}}$^{attn}$ (L6T3N4) 
        & 75.13 & 15.14M \\
        \emph{\textbf{WONN}}$^{attn}$ (L6T6N2) 
        & 76.40 & 12.04M \\
        \emph{\textbf{WONN}}$^{attn}$ (L9T3N2) 
        & 76.77 & 18.28M \\
        \midrule
        \multicolumn{3}{l}{\textit{$S(\theta_i)\&I(\theta_i)$ as trigonometric functions, Ch=256, local (conv) and globle (attn) coupling}}\\
        \emph{\textbf{WONN}}$^{attn}$ (L1T3N2) 
        & 66.07 & 1.62M \\
        \emph{\textbf{WONN}}$^{attn}$ (L6T1N2) 
        & 74.58 & 11.86M \\
        \emph{\textbf{WONN}}$^{attn}$ (L6T3N2) 
        & 76.17 & 11.86M \\
        \emph{\textbf{WONN}}$^{attn}$ (L6T3N1) 
        & 76.94 & 11.79M \\
        \emph{\textbf{WONN}}$^{conv}$ (L6T3N2) 
        & 76.27 & 18.54M \\
        \emph{\textbf{WONN}}$^{attn}$ (L6T3N4) 
        & 76.24 & 12.67M \\
        
        \bottomrule
    \end{tabular}
\end{minipage}
\FloatBarrier




\noindent\textbf{Impact of Patch Size.}
Table~\ref{tab:imagenet100_addition} studies the input patch size on ImageNet-100.
For both MLPs-based and trigonometric interaction functions, a smaller patch size generally improves performance, suggesting that finer spatial tokenization might provides more effective oscillatory representations.

\noindent\textbf{Impact of Interaction Parameterization.}
We compare two choices for the sensitivity and influence functions, namely learnable MLPs and fixed trigonometric functions.
Both variants achieve strong performance, indicating that the benefit of \emph{WONN} mainly comes from the oscillatory dynamical structure rather than a specific parameterization. The trigonometric form provides a stronger oscillator-inspired inductive bias and tends to outperfrom in small datasets, while the learnable MLP parameterization offers greater flexibility, which may be beneficial for scaling to larger and more complex datasets.

\noindent\textbf{Impact of Depth (L).}
Increasing the number of Winfree dynamics layers generally improves performance. For example, moving from shallow layers to \(L=6\) substantially improves CIFAR-100 accuracy, showing that stacked dynamical refinement is important for representation learning.

\noindent\textbf{Impact of Dynamics Steps (T).}
Increasing the number of recurrent dynamics steps also tends to improve performance, but at the cost of additional computation.
This suggests a trade-off between the evolution horizon of synchronization dynamics and computational efficiency.

\noindent\textbf{Impact of Group Size (N).}
The effect of group size depends on the interaction form.
MLP-based interactions perform best at \(N=2\), while trigonometric interactions benefit from a different grouping choice, with \(N=1\) outperforming \(N=2\) among grouped variants.
This suggests that selecting an appropriate group size can enhance WONN performance by matching the interaction scale to the parameterization of the interaction function.

\noindent\textbf{Impact of Coupling Type (conv vs. attn).}
We compare global attentive coupling with local convolutional coupling under both MLP-based and trigonometric interactions.
Although convolutional coupling achieves comparable accuracy, it introduces substantially more parameters in our implementation.
Attentive coupling therefore offers a more efficient mechanism for coordinating oscillators, and its better parameter efficiency may make it more favorable for scaling WONN to larger and more complex datasets.

\subsection{Fixed and Learnable Components}
\label{app:fixed_learnable}

Table~\ref{tab:fixed_learnable} distinguishes fixed and
learnable components of \emph{WONN} under the trigonometric
and MLP-based parameterizations.

\begin{table}[htbp]
    \centering
    \caption{Fixed and learnable components of \emph{WONN}.
    ``Fixed'' denotes operations without trainable parameters
    or manually specified hyperparameters.}
    \label{tab:fixed_learnable}
    \small
    \setlength{\tabcolsep}{6pt}
    \begin{tabular}{lcc}
        \toprule
        \textbf{Component}
        & \textbf{Trigonometric}
        & \textbf{MLP-based} \\
        \midrule
        Point-wise sensitivity $S(\theta)$
        & Fixed $\cos\theta$
        & Learnable MLP \\

        Point-wise influence $I(\theta)$
        & Fixed $\sin\theta$
        & Learnable MLP \\

        Coupling (attention or convolution)
        & Learnable
        & Learnable \\

        Layer-transition mappings
        & Learnable
        & Learnable \\

        Input embedding and output head
        & Learnable
        & Learnable \\

        Hyperparameters $\gamma$, $L$, $T_l$, and $N$
        & Fixed
        & Fixed \\
        \bottomrule
    \end{tabular}
\end{table}
\FloatBarrier

\subsection{Solving Maze-hard} 

Empirically, we observe that the interaction energy $E_{int}$ consistently decreases as the number of time steps $T$ increases, serving as a reliable metric for the degree of synchronization (as detailed in Sec.~\ref{sec:winfree_property}). Leveraging this property, we introduce the energy-based voting mechanism during inference to enhance model performance. Specifically, we appropriately extend the evaluation time steps $T$ and perform $K$ independent trials with different random initializations of initial phases. The prediction corresponding to the lowest final interaction energy $E_{int}$ is ultimately selected as the optimal solution, which empirically boosts the performance.

\FloatBarrier
\noindent
\begin{minipage}{\columnwidth}
    \centering
    \captionsetup{hypcap=false}
    \captionof{table}{Solving Maze-hard results}
    \label{tab:maze_addition}
    \small
    \setlength{\tabcolsep}{4pt}
    \begin{tabular}{lcc}
        \toprule
        \textbf{Models} & \textbf{Accuracy (\%)} & \textbf{$\#$ Parameters} \\
        \midrule
        \multicolumn{3}{l}{\textit{Other synchrony-based model}} \\
        \textbf{\emph{AKOrN}}$^{attn}$(L1T16)   & 32.3 & 1M \\
        \textbf{\emph{AKOrN}}$^{attn}$(L1T24)   & 36.2 & 1M \\
        \midrule
        \multicolumn{3}{l}{\textit{$S(\theta_i)$ \& $I(\theta_i)$ as trigonometric functions, L=1, T=24}} \\
        \emph{\textbf{WONN}}      & 76.2 & \textbf{0.396M} \\
        \emph{\textbf{WONN}} ($T_{eval}=25$, K=32)      & 80.1 & \textbf{0.396M} \\
        \bottomrule
    \end{tabular}
\end{minipage}

\subsection{Computational Cost and Memory Usage}
\label{app:compute_cost}

We report additional measurements of training latency,
single-sample inference latency, and peak GPU memory on
Sudoku and Maze-hard. All measurements are conducted on
a single NVIDIA H100 using the same numerical precision
and timing protocol as the CIFAR-10 measurements in
Sec.~\ref{sec:compute_cost}. Training latency is measured
with batch size 32.

\noindent\textbf{Sudoku.}
We compare \emph{WONN} with the corresponding \emph{AKOrN}
reasoning model in Table~\ref{tab:sudoku_cost}.

\begin{table}[htbp]
    \centering
    \caption{Training and inference costs on Sudoku.}
    \label{tab:sudoku_cost}
    \small
    \setlength{\tabcolsep}{4pt}
    \begin{tabular}{lccccc}
        \toprule
        \multirow{2}{*}{\textbf{Model}}
        & \multirow{2}{*}{\textbf{Params.}}
        & \multicolumn{2}{c}{\textbf{Latency (ms)}}
        & \multicolumn{2}{c}{\textbf{Peak memory (GiB)}} \\
        \cmidrule(lr){3-4} \cmidrule(lr){5-6}
        & & Train/batch & Infer/sample & Train & Infer \\
        \midrule
        \emph{WONN}
        & 1.583M
        & $ \textbf{28.523} \pm 0.106$
        & $ \textbf{1.820} \pm 0.018$
        & 0.336 & 0.042 \\
        \emph{AKOrN}
        & 2.982M
        & $37.272 \pm 0.044$
        & $2.247 \pm 0.017$
        & 0.780 & 0.020 \\
        \bottomrule
    \end{tabular}
\end{table}
\FloatBarrier

On Sudoku, \emph{WONN} uses fewer parameters and achieves
lower training and inference latency and lower training
memory than \emph{AKOrN}.

\noindent\textbf{Maze-hard.}
For \emph{WONN} and \emph{AKOrN}, the reported inference
latency corresponds to a single trajectory ($K=1$).
\emph{HRM} and \emph{TRM} use their standard deterministic inference procedures. Table~\ref{tab:maze_cost} reports the results.

\begin{table}[htbp]
    \centering
    \caption{Training and inference costs on Maze-hard.}
    \label{tab:maze_cost}
    \small
    \setlength{\tabcolsep}{4pt}
    \begin{tabular}{lccccc}
        \toprule
        \multirow{2}{*}{\textbf{Model}}
        & \multirow{2}{*}{\textbf{Params.}}
        & \multicolumn{2}{c}{\textbf{Latency (ms)}}
        & \multicolumn{2}{c}{\textbf{Peak memory (GiB)}} \\
        \cmidrule(lr){3-4} \cmidrule(lr){5-6}
        & & Train/batch & Infer/sample & Train & Infer \\
        \midrule
        \emph{WONN}
        & 0.396M
        & $ \textbf{40.683} \pm 0.494$
        & $ \textbf{3.099} \pm 0.067$
        & 2.021 & 0.036 \\
        \emph{AKOrN}
        & 0.957M
        & $62.927 \pm 0.003$
        & $ \textbf{2.990} \pm 0.035$
        & 6.146 & 0.011 \\
        \emph{HRM}$^{\dagger}$
        & 27.270M
        & $51.347 \pm 0.205$
        & $50.279 \pm 1.233$
        & 5.081 & 0.148 \\
        \emph{TRM}$^{\dagger}$
        & 6.823M
        & $43.214 \pm 0.152$
        & $48.093 \pm 0.223$
        & 5.706 & 0.087 \\
        \bottomrule
    \end{tabular}

    \smallskip
    \begin{minipage}{\linewidth}
        \footnotesize
        $\dagger$ The reported \emph{HRM} and \emph{TRM} training latencies
        correspond to only one Adaptive Computation Time
        (ACT) step. A complete update executes a data-dependent number of ACT steps, up to 16, whereas the \emph{WONN} and \emph{AKOrN} measurements correspond to complete training updates.
    \end{minipage}
\end{table}
\FloatBarrier

On Maze-hard, \emph{WONN} completes a full training update
faster than \emph{AKOrN} and even faster than a single ACT
step of \emph{HRM} or \emph{TRM}. It also has the lowest measured training
memory usage among these models. At inference with $K=1$, \emph{WONN} is comparable to \emph{AKOrN} and approximately \textbf{ \bm{$16.2\times$} faster than \emph{HRM} and \bm{$15.5\times$} faster than \emph{TRM}}.

\noindent\textbf{Energy-voting cost.}
For the $K=32$ energy-voting setting used to obtain
80.1\% Maze-hard accuracy, we additionally measure serial
and parallel execution, as reported in
Table~\ref{tab:energy_voting_cost}.

\begin{table}[htbp]
    \centering
    \caption{Inference costs of energy voting on Maze-hard
    with $K=32$ trajectories. Speedup is relative to serial
    execution.}
    \label{tab:energy_voting_cost}
    \small
    \setlength{\tabcolsep}{6pt}
    \begin{tabular}{lccc}
        \toprule
        \textbf{Execution mode}
        & \textbf{Latency (ms/maze)}
        & \textbf{Speedup}
        & \textbf{Peak memory (GiB)} \\
        \midrule
        Serial
        & $97.197 \pm 1.087$
        & $1.000\times$
        & 0.038 \\
        Parallel
        & $ \textbf{13.158} \pm 0.132$
        & $7.387\times$
        & 0.104 \\
        \bottomrule
    \end{tabular}
\end{table}
\FloatBarrier

Parallel execution reduces latency by a factor of
$7.387$. Even with $K=32$ trajectories, parallel
\emph{WONN} inference remains approximately \textbf{\bm{$3.8\times$} faster than \emph{HRM} and \bm{$3.7\times$} faster than \emph{TRM}} under
their standard inference procedures.

\section{Energy Structure and Topological Obstruction}
\label{app:lyapunov}

We clarify the energy structure of the trigonometric Winfree dynamics used in \emph{WONN}.
For simplicity, we analyze the continuous-time phase dynamics
\begin{equation}
\dot{\theta}_i
=
\omega_i
+
\gamma \cos\theta_i
\sum_j c_{ij}\sin\theta_j,
\qquad
\theta_i\in\mathbb{T}:=\mathbb{R}/(2\pi\mathbb{Z}),
\label{eq:appendix-trig-winfree}
\end{equation}
where $\omega_i$ is the natural frequency, $\gamma>0$ is the interaction strength, and $c_{ij}$ is the coupling coefficient from oscillator $j$ to oscillator $i$.

\paragraph{Absence of a global potential in the presence of natural frequencies.}
We first show that the natural-frequency term prevents the full dynamics from being written as a globally defined gradient flow on the torus.
Consider the pure frequency dynamics
\begin{equation}
\dot{\theta}_i=\omega_i .
\end{equation}
If this vector field were the negative gradient of a globally defined scalar potential 
$E:\mathbb{T}^d\to\mathbb{R}$, then in local phase coordinates it would satisfy
\begin{equation}
-\frac{\partial E}{\partial \theta_i}
=
\omega_i,
\qquad\text{or equivalently}\qquad
\frac{\partial E}{\partial \theta_i}
=
-\omega_i .
\end{equation}
Locally, this implies
\begin{equation}
E(\Theta)
=
-\sum_i \omega_i\theta_i+\mathrm{const}.
\end{equation}
However, $\theta_i$ is not a globally single-valued function on $\mathbb{T}$.
Since $\theta_i$ and $\theta_i+2\pi$ represent the same point on the circle, a globally defined potential must be invariant under the identification
\[
\theta_i\sim \theta_i+2\pi .
\]
But the local potential changes by
\begin{equation}
E(\Theta+2\pi e_i)-E(\Theta)
=
-2\pi\omega_i .
\end{equation}
Therefore, when $\omega_i\neq 0$, this local potential is not single-valued on the torus.
Equivalently, the constant frequency drift has nonzero circulation along the fundamental cycle of the $i$-th circle:
\begin{equation}
\oint \omega_i\, d\theta_i
=
2\pi\omega_i
\neq 0 .
\end{equation}
Since the gradient of a globally defined scalar potential must have zero circulation along every closed loop, the natural-frequency term is a topological obstruction to a global potential on $\mathbb{T}^d$.

\paragraph{Interaction energy.}
Although the full dynamics do not admit a globally defined potential in the presence of natural frequencies, the trigonometric interaction term admits a natural energy under symmetric coupling.
Assume that the coupling matrix is fixed and symmetric:
\begin{equation}
c_{ij}=c_{ji}.
\end{equation}
Define the interaction energy
\begin{equation}
E_{\mathrm{int}}(\Theta)
=
-\frac{1}{2}
\sum_{i,j}
c_{ij}\sin\theta_i\sin\theta_j .
\label{eq:eint}
\end{equation}
Then
\begin{equation}
\frac{\partial E_{\mathrm{int}}}{\partial\theta_i}
=
-\cos\theta_i
\sum_j c_{ij}\sin\theta_j .
\label{eq:eint-gradient}
\end{equation}
Therefore, the trigonometric interaction term can be written as
\begin{equation}
\cos\theta_i
\sum_j c_{ij}\sin\theta_j
=
-
\frac{\partial E_{\mathrm{int}}}{\partial\theta_i}.
\end{equation}
Substituting this into Eq.~\eqref{eq:appendix-trig-winfree}, the full dynamics can be decomposed as
\begin{equation}
\dot{\theta}_i
=
\omega_i
-
\gamma
\frac{\partial E_{\mathrm{int}}}{\partial\theta_i}.
\label{eq:drift-plus-gradient}
\end{equation}
Thus, the system is a drift-plus-gradient system rather than a pure gradient flow.

The same structure can also be expressed geometrically on the embedded circle.
Let
\begin{equation}
x_i
=
\begin{pmatrix}
\cos\theta_i\\
\sin\theta_i
\end{pmatrix}
\in S^1,
\qquad
T_i
=
\frac{\partial x_i}{\partial \theta_i}
=
\begin{pmatrix}
-\sin\theta_i\\
\cos\theta_i
\end{pmatrix}.
\end{equation}
Here $T_i$ is the unit tangent vector at $x_i$, satisfying $x_i^\top T_i=0$.
Since $\dot{x}_i=T_i\dot{\theta}_i$, Eq.~\eqref{eq:drift-plus-gradient} becomes
\begin{equation}
\dot{x}_i
=
\omega_i T_i
-
\gamma\, \mathrm{grad}^{S^1}_{x_i}E_{\mathrm{int}} .
\end{equation}
The first term is the natural-frequency drift along the circle, while the second term is the Riemannian gradient flow induced by the interaction energy.

\paragraph{Why full dynamics does not admit a Lyapunov function.}
Although $E_{\mathrm{int}}$ captures the interaction part of the dynamics, it is not in general a Lyapunov function for the full system when $\omega\neq 0$.
Indeed, along trajectories of Eq.~\eqref{eq:drift-plus-gradient},
\begin{align}
\frac{dE_{\mathrm{int}}}{dt}
&=
\sum_i
\frac{\partial E_{\mathrm{int}}}{\partial\theta_i}
\dot{\theta}_i \\
&=
\sum_i
\omega_i
\frac{\partial E_{\mathrm{int}}}{\partial\theta_i}
-
\gamma
\sum_i
\left(
\frac{\partial E_{\mathrm{int}}}{\partial\theta_i}
\right)^2 .
\end{align}
Equivalently,
\begin{equation}
\frac{dE_{\mathrm{int}}}{dt}
=
\left\langle
\nabla_\Theta E_{\mathrm{int}},
\omega
\right\rangle
-
\gamma
\left\|
\nabla_\Theta E_{\mathrm{int}}
\right\|_2^2 .
\end{equation}
The second term is non-positive, but the first term is not sign-definite.
Consequently, $E_{\mathrm{int}}$ is not guaranteed to decrease along the full dynamics, and therefore it is not a Lyapunov function for the system with nonzero natural frequencies.

\paragraph{Lyapunov structure in the zero-frequency symmetric-coupling regime.}
If the natural-frequency term is removed, \emph{i.e.,}
\begin{equation}
\omega_i=0
\qquad
\text{for all } i,
\end{equation}
and the coupling matrix is fixed and symmetric, then Eq.~\eqref{eq:drift-plus-gradient} reduces to
\begin{equation}
\dot{\theta}_i
=
-
\gamma
\frac{\partial E_{\mathrm{int}}}{\partial\theta_i}.
\end{equation}
Thus the dynamics are a gradient flow on the torus:
\begin{equation}
\dot{\Theta}
=
-
\gamma
\nabla_\Theta E_{\mathrm{int}}(\Theta).
\end{equation}
In this case,
\begin{align}
\frac{dE_{\mathrm{int}}}{dt}
&=
\sum_i
\frac{\partial E_{\mathrm{int}}}{\partial\theta_i}
\dot{\theta}_i \\
&=
-
\gamma
\sum_i
\left(
\frac{\partial E_{\mathrm{int}}}{\partial\theta_i}
\right)^2 \\
&\leq 0 .
\end{align}
Therefore, in the zero-frequency symmetric-coupling regime, $E_{\mathrm{int}}$ is a Lyapunov function for the trigonometric interaction dynamics.
In particular, isolated local minima of $E_{\mathrm{int}}$ correspond to Lyapunov-stable synchronized states.

\paragraph{Implication for \emph{WONN}.}
The full \emph{WONN} architecture contains frequency states, discrete-time updates, layer transitions and non-symmetric or state-dependent interactions.
Therefore, the Lyapunov structure above should not be interpreted as a global stability theorem for the full model. Nonetheless, $E_{\mathrm{int}}$ provides a principled interaction energy for the trigonometric Winfree core and serves as a practical diagnostic of phase alignment. In Maze-hard (Sec.~\ref{sec:Maze_hard}), we use this diagnostic for energy-based test-time selection among multiple stochastic phase trajectories.

\section{Additional Visualization}

\subsection{Additional Visualization on Image Recognition}
\label{app:Additional Visualization on Image Recognition}

\begin{figure}[t]
    \centering
    \begin{minipage}{0.48\columnwidth}
        \centering
        \includegraphics[width=\linewidth]{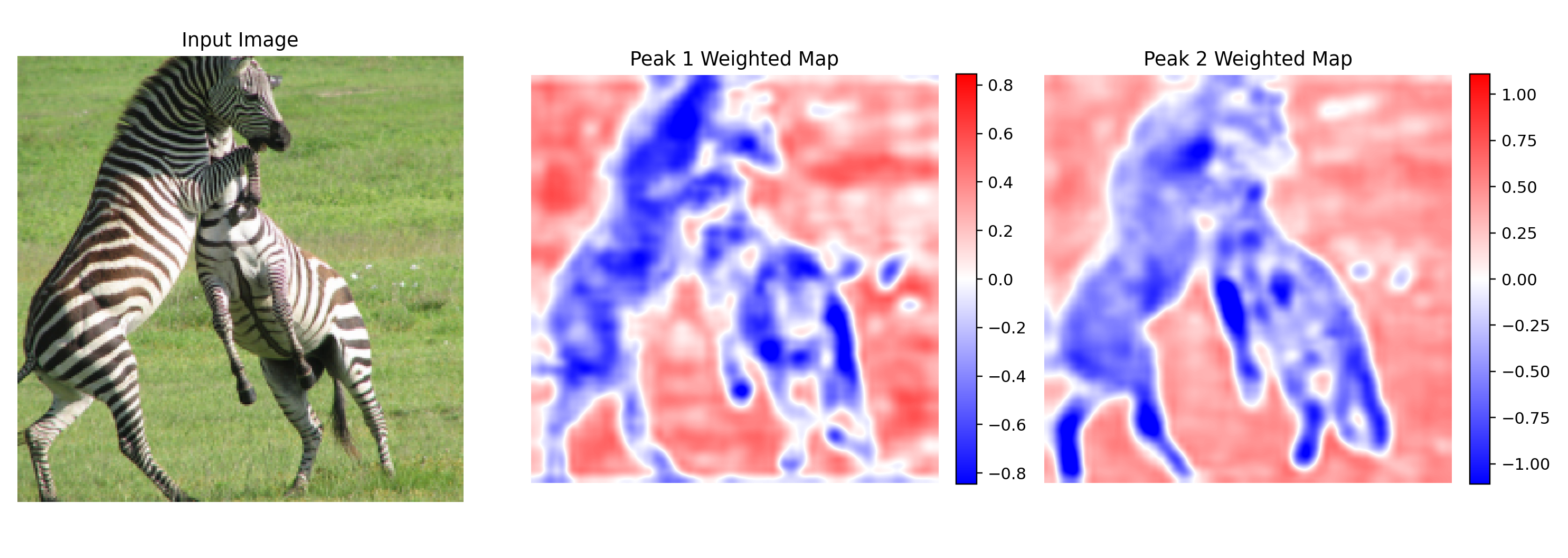}
    \end{minipage}
    \hfill
    \begin{minipage}{0.48\columnwidth}
        \centering
        \includegraphics[width=\linewidth]{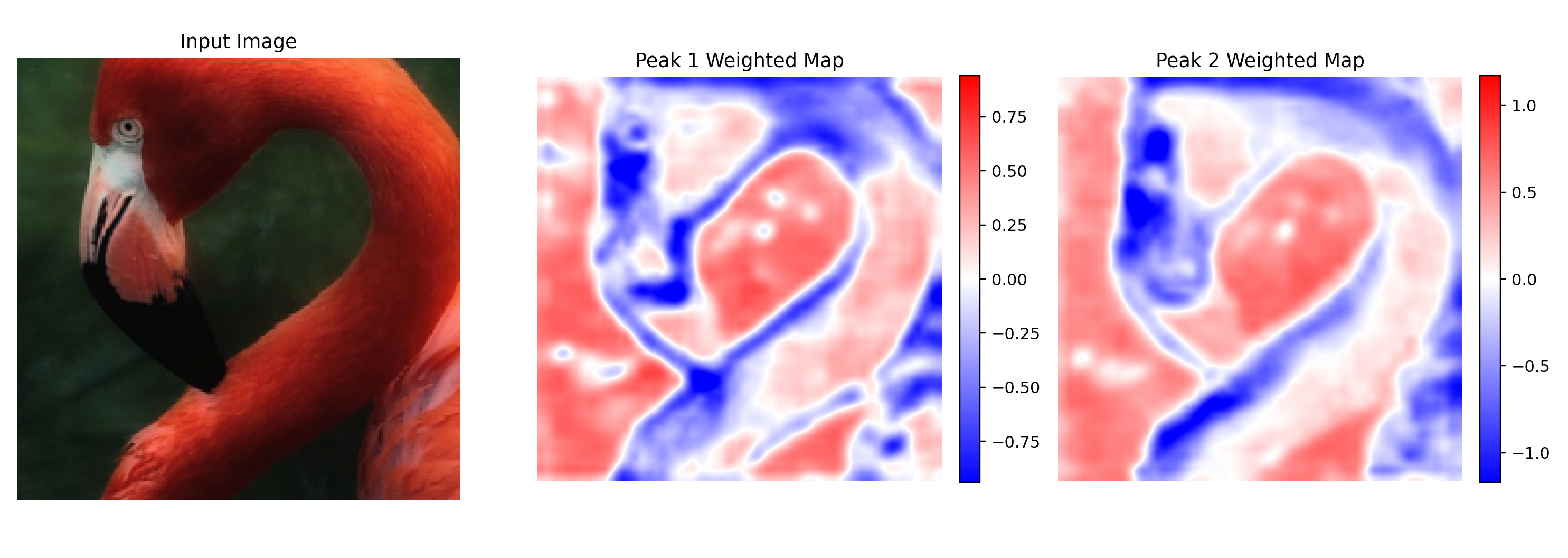}
    \end{minipage}
    \vspace{0.5em}
    \begin{minipage}{0.48\columnwidth}
        \centering
        \includegraphics[width=\linewidth]{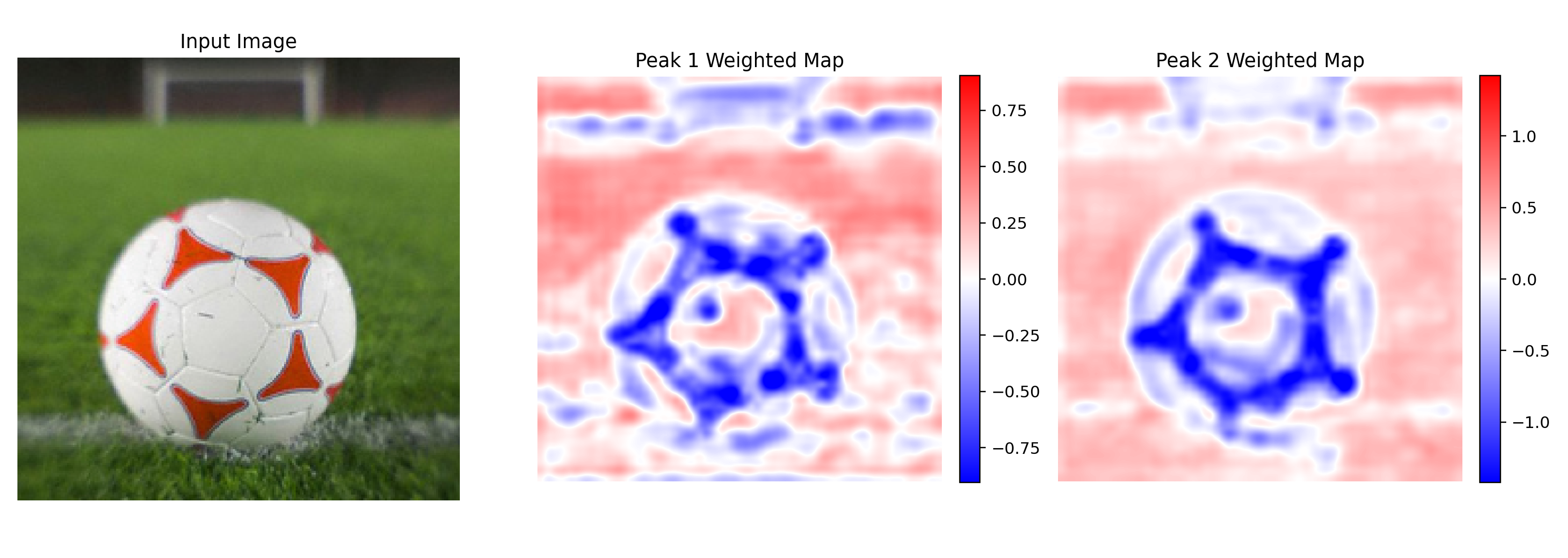}
    \end{minipage}
    \hfill
    \begin{minipage}{0.48\columnwidth}
        \centering
        \includegraphics[width=\linewidth]{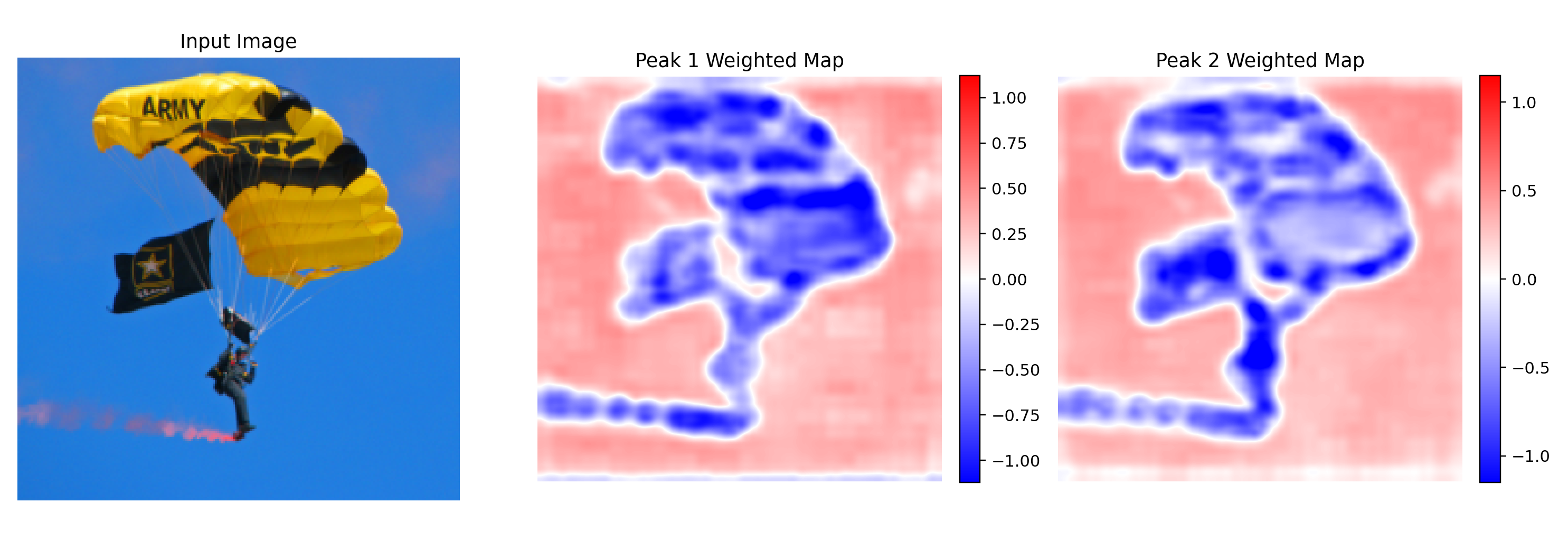}
    \end{minipage}
    \caption{
    Additional qualitative visualizations of two-peak distributions on image recognition.
    }
    \label{fig:additional_image_visualization}
\end{figure}

Here we provide additional qualitative visualizations for image recognition. As shown in Fig.~\ref{fig:additional_image_visualization}, these examples exhibit similar patterns to those discussed in Sec.~\ref{sec:image_recognition}, further supporting the observation that the phase modes of \emph{WONN} capture complementary visual structures.


\begin{figure}[t]
    \centering
    \includegraphics[width=\linewidth]{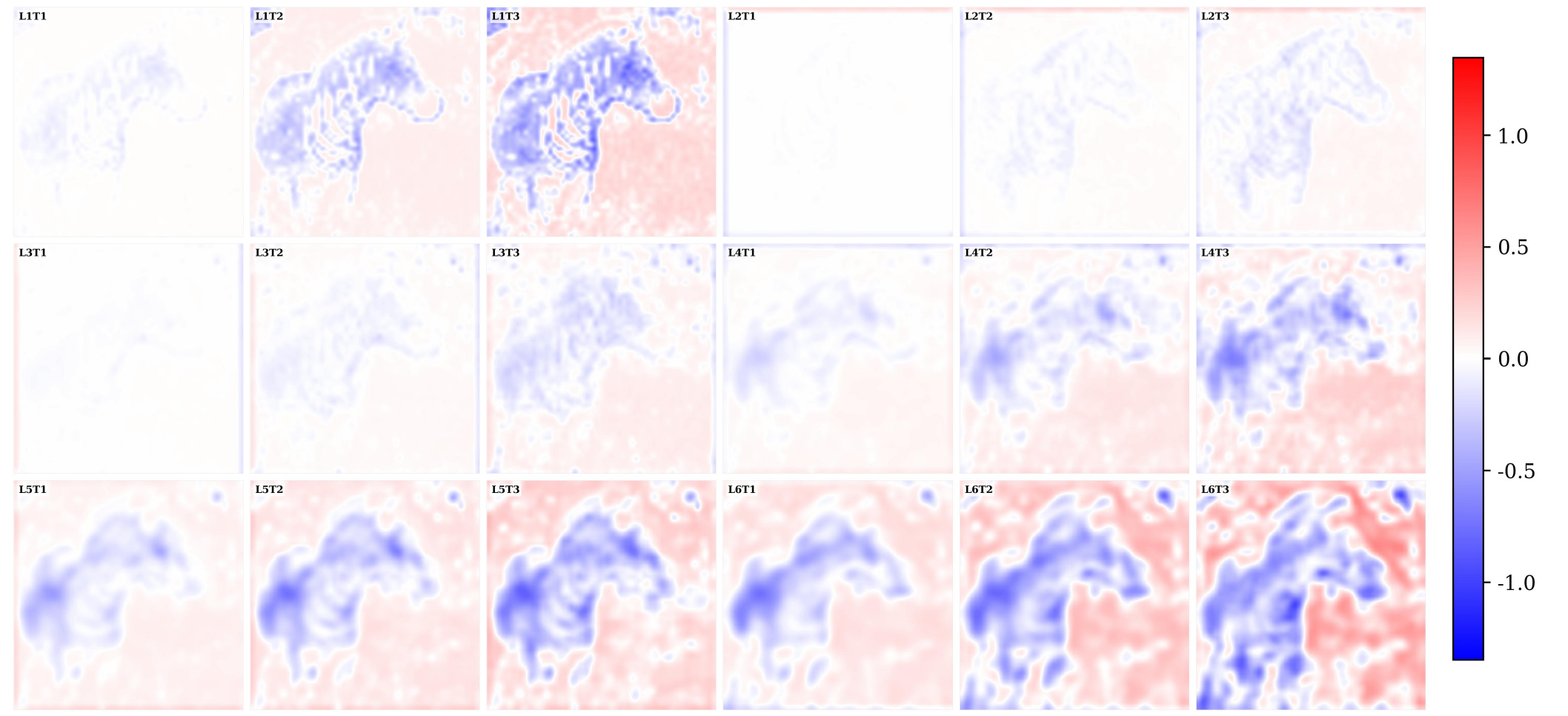}

    \vspace{0.8em}

    \includegraphics[width=\linewidth]{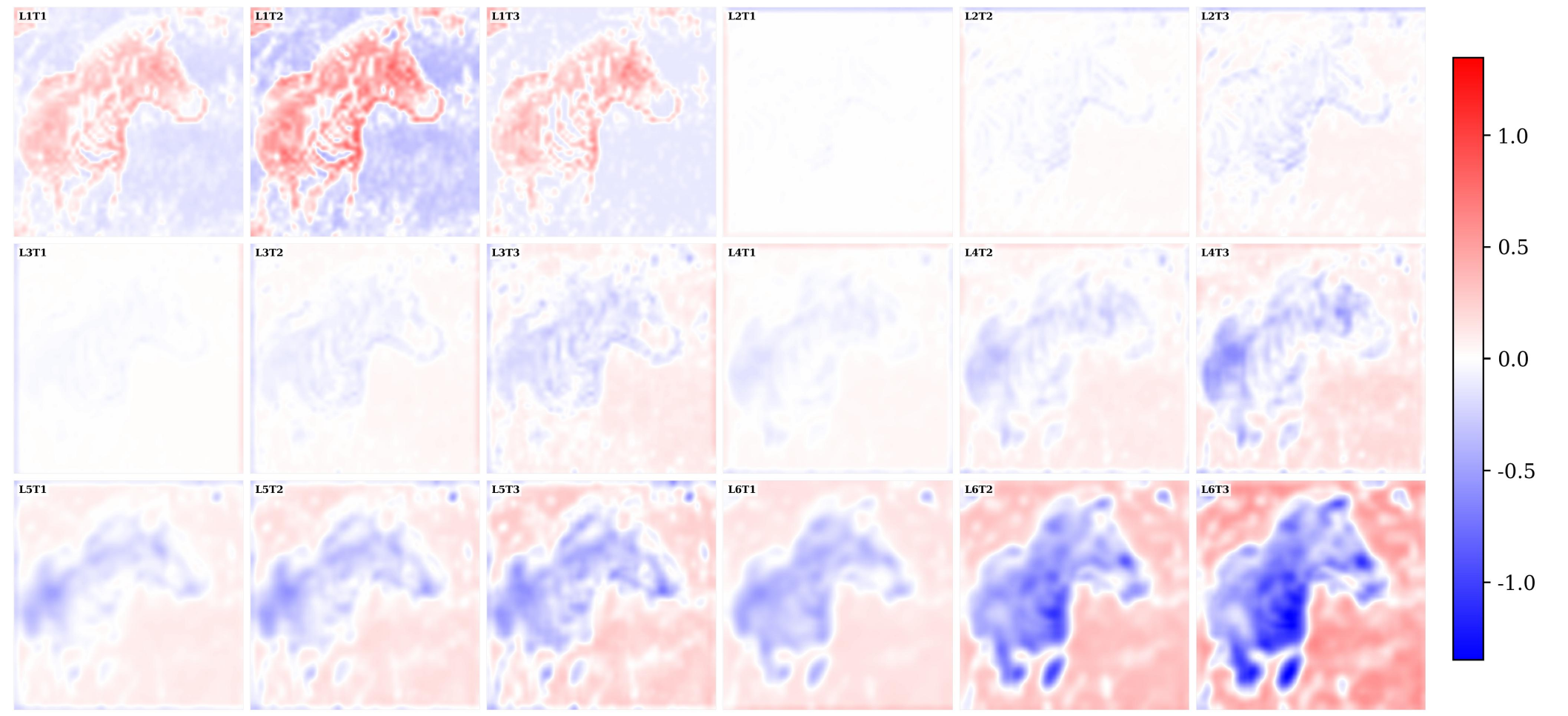}

    \caption{
    Layer-wise evolution of the weighted maps associated with the two dominant phase peaks in \emph{WONN} on image recognition. \textbf{Top}: weighted map corresponding to the first dominant phase peak. \textbf{Bottom}: weighted map corresponding to the second dominant phase peak. Here \(L\) denotes the layer index and \(T\) denotes the Winfree dynamics step within that layer. Panels are arranged according to the actual forward trajectory, from \(L1T1\) to \(L6T3\). Across layers, the phase-weighted responses are progressively refreshed and reorganized, evolving from weak local activations toward more coherent global semantic structures through the recurrent synchronization dynamics.
    }
    \label{fig:layer_dynamics_image_recognition}
\end{figure}

In addition to the static weighted maps shown in Sec.~\ref{sec:image_recognition}, we further visualize their layer-wise evolution in Fig.~\ref{fig:layer_dynamics_image_recognition}. The figures show how the weighted maps associated with the two dominant phase peaks evolve across recurrent Winfree dynamics steps and network depth. We observe that the phase-weighted responses are repeatedly refreshed at each layer: early layers mainly capture weak and local visual structures, while deeper layers progressively organize these responses into more coherent global semantic patterns. This behavior suggests that the recurrent synchronization dynamics of \emph{WONN} gradually refine and reorganize visual representations over depth and time.

\newpage
\subsection{Additional Visualization on Maze-hard}
\label{sec:additional_visualization_maze_hard}

This subsection presents additional Maze-hard examples for qualitative analysis. 

\begin{wrapfigure}[20]{r}{0.5\columnwidth}
    \vspace{-1.0\baselineskip}
    \centering
    \includegraphics[width=\linewidth]{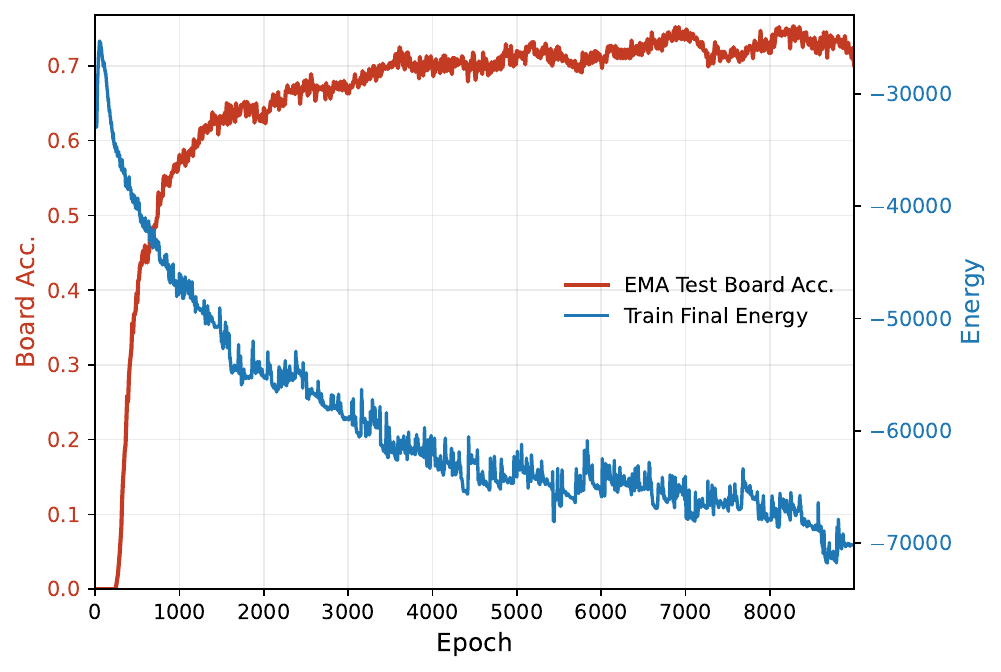}
    \caption{
    Accuracy--energy dynamics on Maze-hard. 
    We report the test accuracy and the corresponding energy over training epochs. 
    After a slight increase in the early stage, the energy gradually decreases as training proceeds, whereas the accuracy exhibits a steady improvement trend.
    }
    \label{fig:maze_energy}
    \vspace{-0.8\baselineskip}
\end{wrapfigure}

Fig.~\ref{fig:maze_energy} illustrates the evolution of accuracy and interaction energy during training on the Maze-hard task. While the energy initially increases in the early stage of optimization, it subsequently decreases as training progresses, accompanied by a steady improvement in accuracy. This inverse correlation suggests that lower interaction energy is associated with more accurate solution states, indicating that the energy provides a useful proxy for solution quality in the learned synchronization dynamics.

\Needspace{20\baselineskip}

\begin{figure}[t]
    \centering

    \includegraphics[width=\columnwidth]{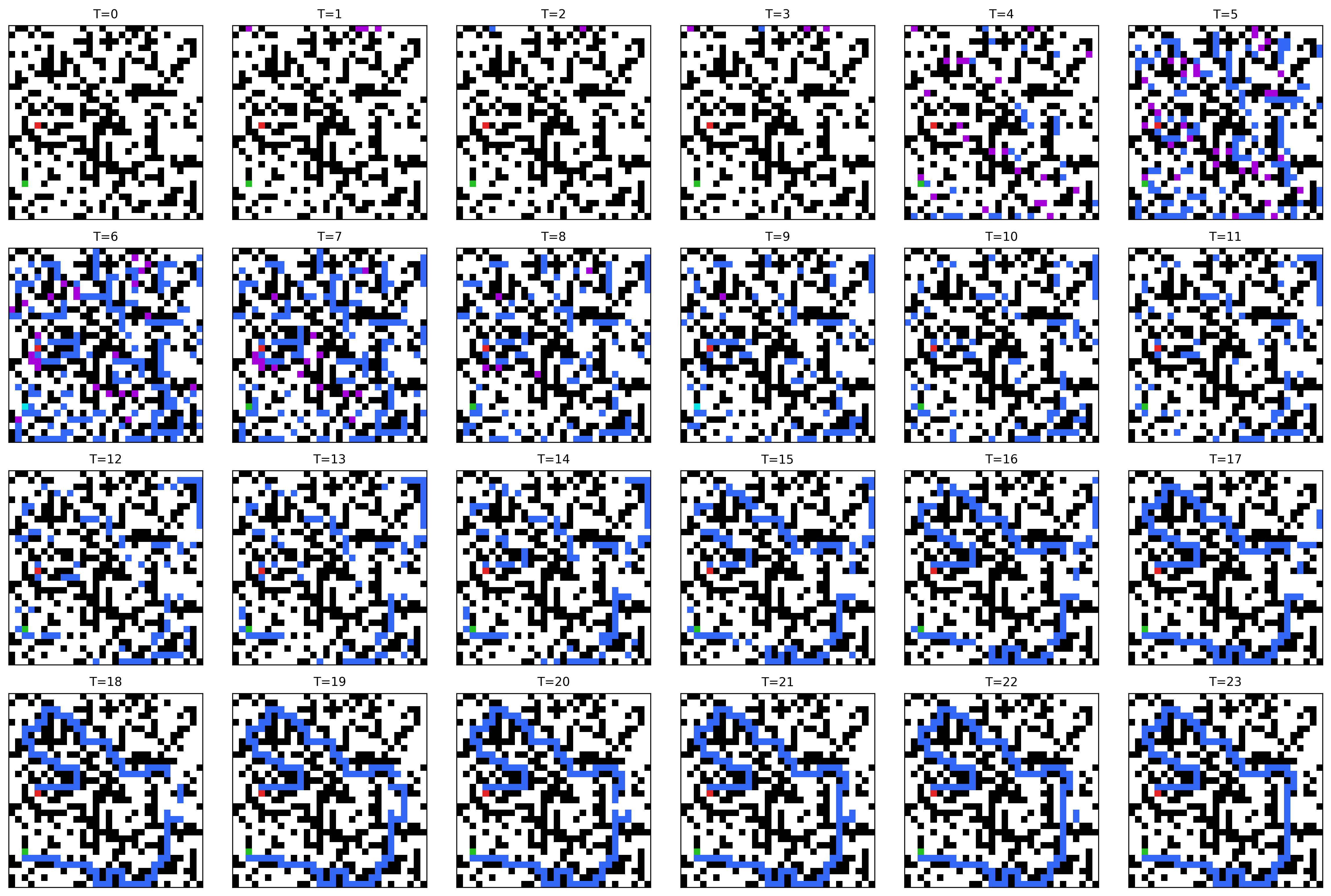}
    \vspace{0.5em}

    \includegraphics[width=\columnwidth]{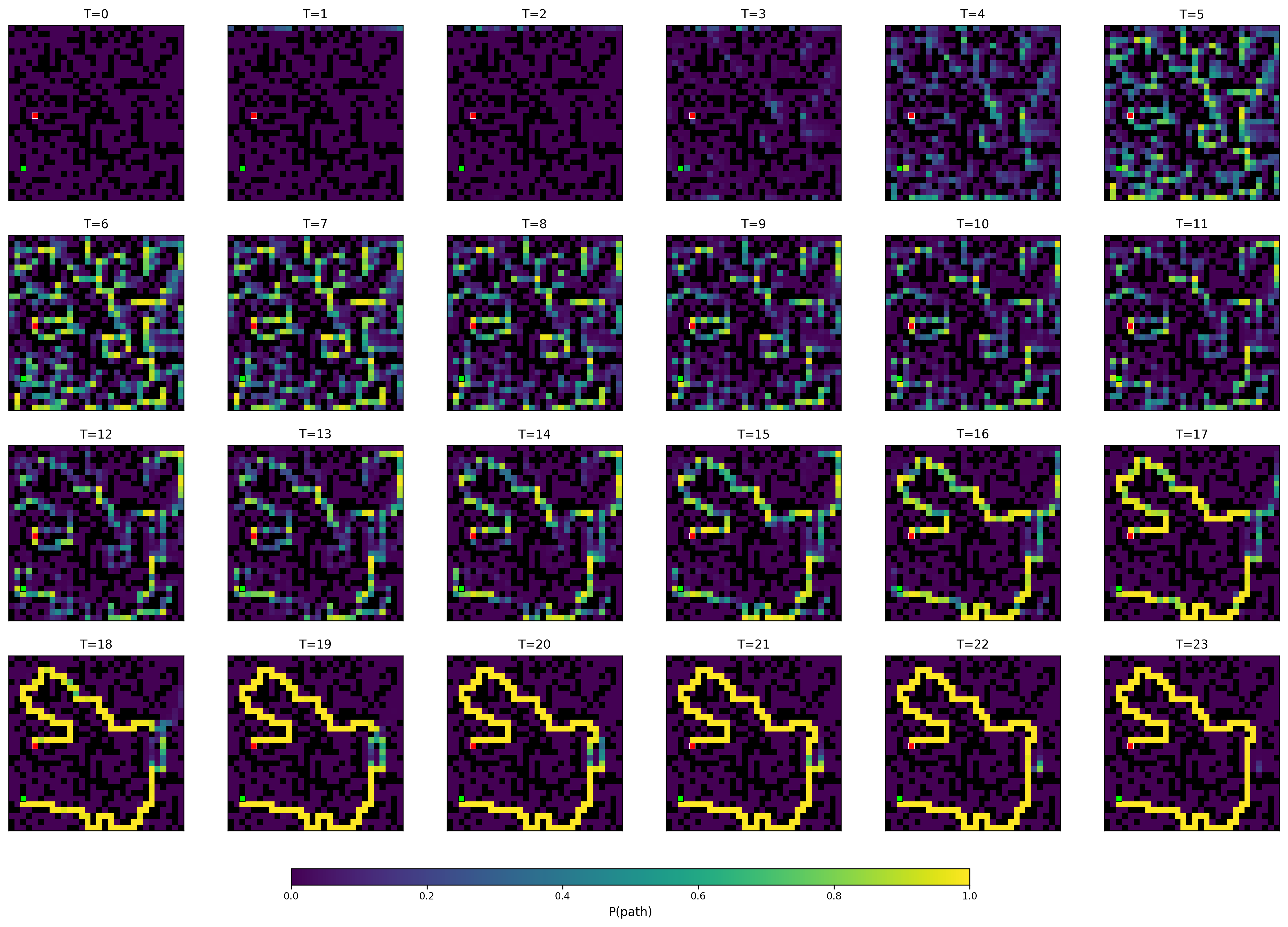}

    \caption{
    Complete temporal evolution of the example shown in Sec.~\ref{sec:Maze_hard}.
    \textbf{Top:} discrete path predictions over time.
    \textbf{Bottom:} predicted path probability heatmaps over time.
    }
    \label{fig:maze_path_1}
\end{figure}

\begin{figure}[t]
    \centering

    \includegraphics[width=\columnwidth]{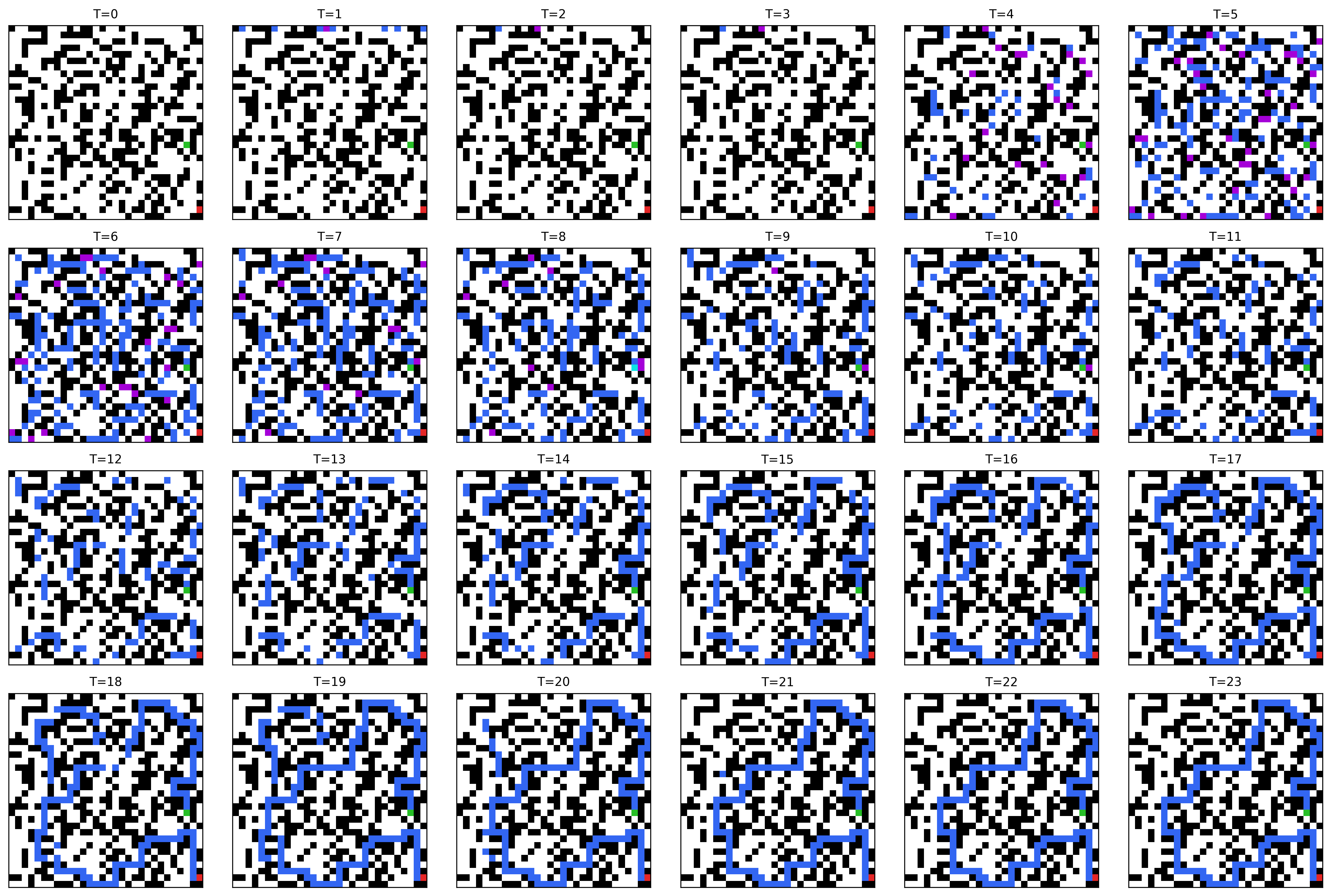}
    \vspace{0.5em}

    \includegraphics[width=\columnwidth]{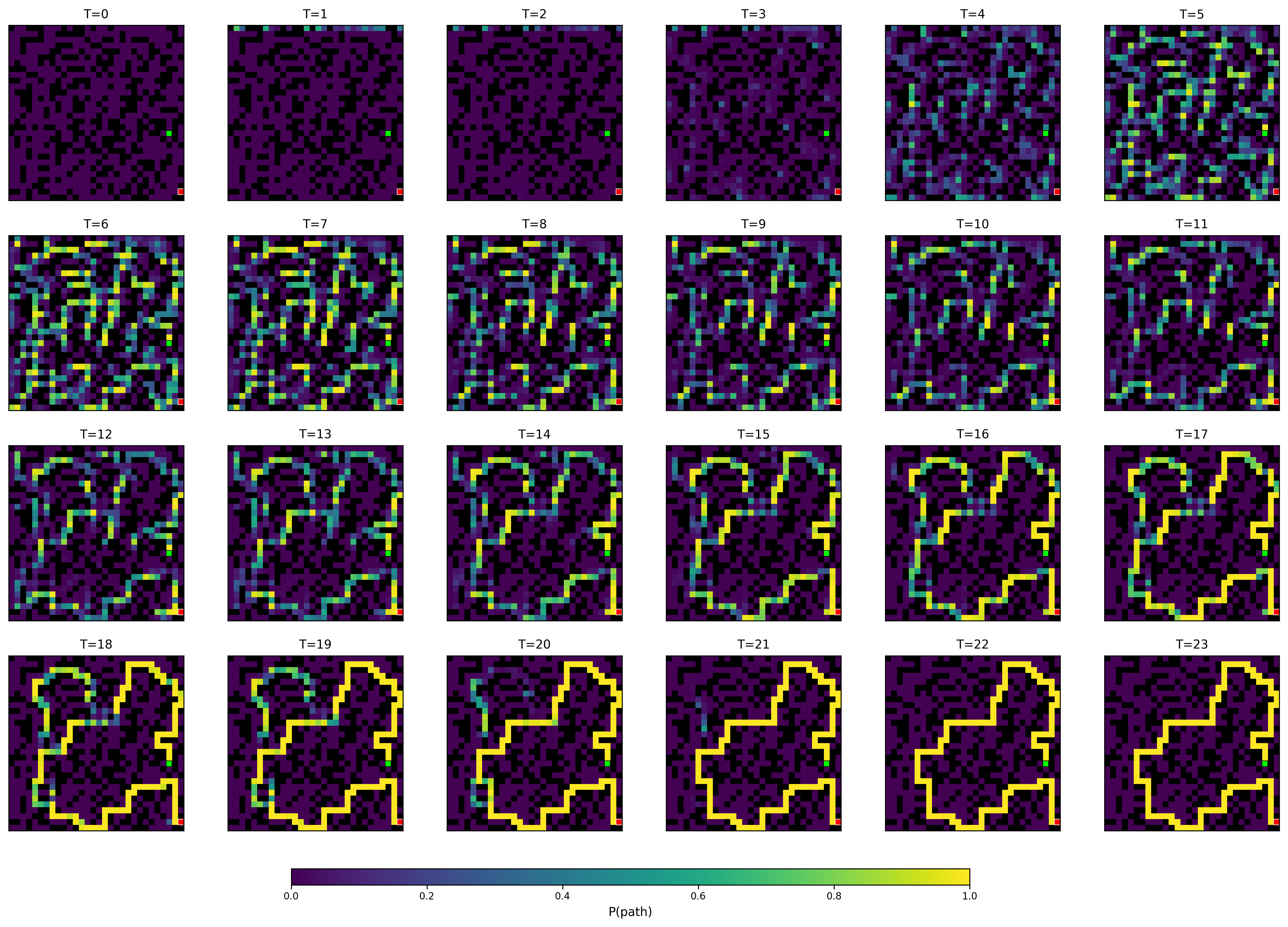}

    \caption{More Maze-hard pathfinding examples. Each pair shows the discrete path prediction and the corresponding path probability heatmap for one maze instance.}
    \label{fig:maze_path_more}
\end{figure}

Fig.~\ref{fig:maze_path_1} shows the complete temporal evolution of the example discussed in Sec.~\ref{sec:Maze_hard}, while Fig.~\ref{fig:maze_path_more} provides several additional instances. In the path prediction maps, black cells represent walls, red and green cells denote the start and goal, respectively, blue cells indicate the predicted path, and purple cells correspond to invalid wall-crossing predictions. Across different examples, we observe a consistent synchronization-driven dynamics: during the early stages, the network activates multiple candidate path fragments simultaneously rather than directly converging to a single solution. As the oscillatory dynamics evolve, these fragments progressively synchronize, compete, and merge into coherent global trajectories, while inconsistent or invalid paths are gradually suppressed. Eventually, the dynamics converge to the shortest valid path connecting the start and goal. These visualizations suggest that synchronization dynamics provide a very natural mechanism for coordinating distributed local candidates into globally consistent reasoning solutions.

\begin{figure}[t]
    \centering
    \includegraphics[width=\columnwidth]{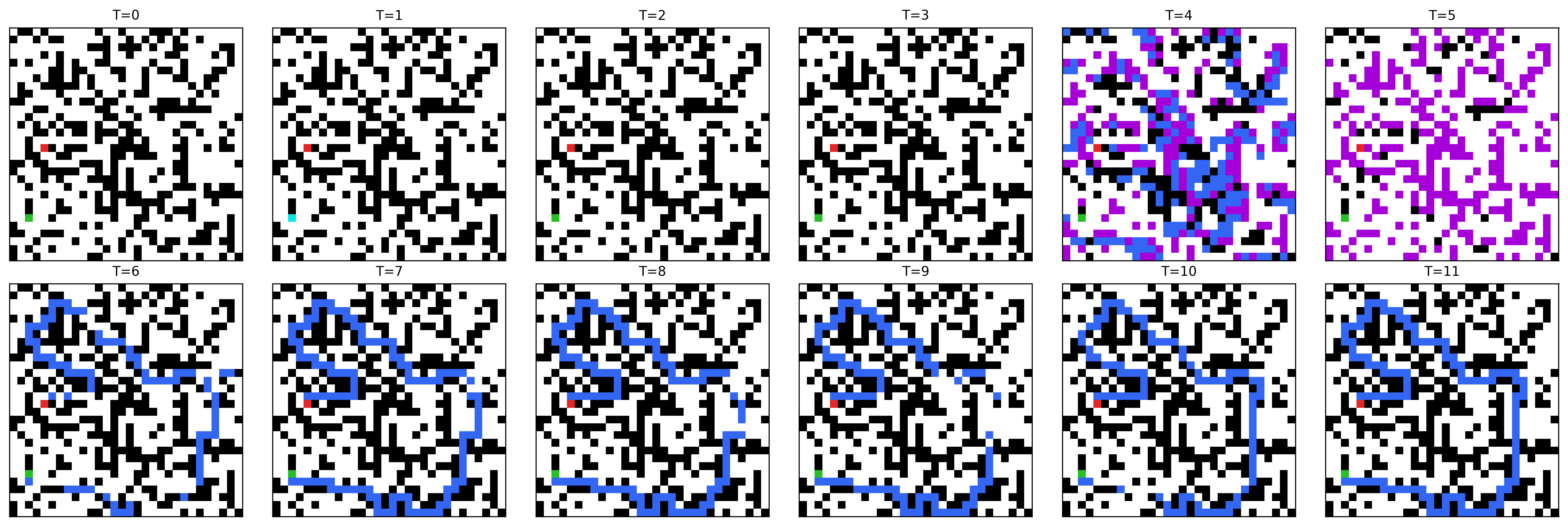}
    \vspace{0.5em}
    \includegraphics[width=\columnwidth]{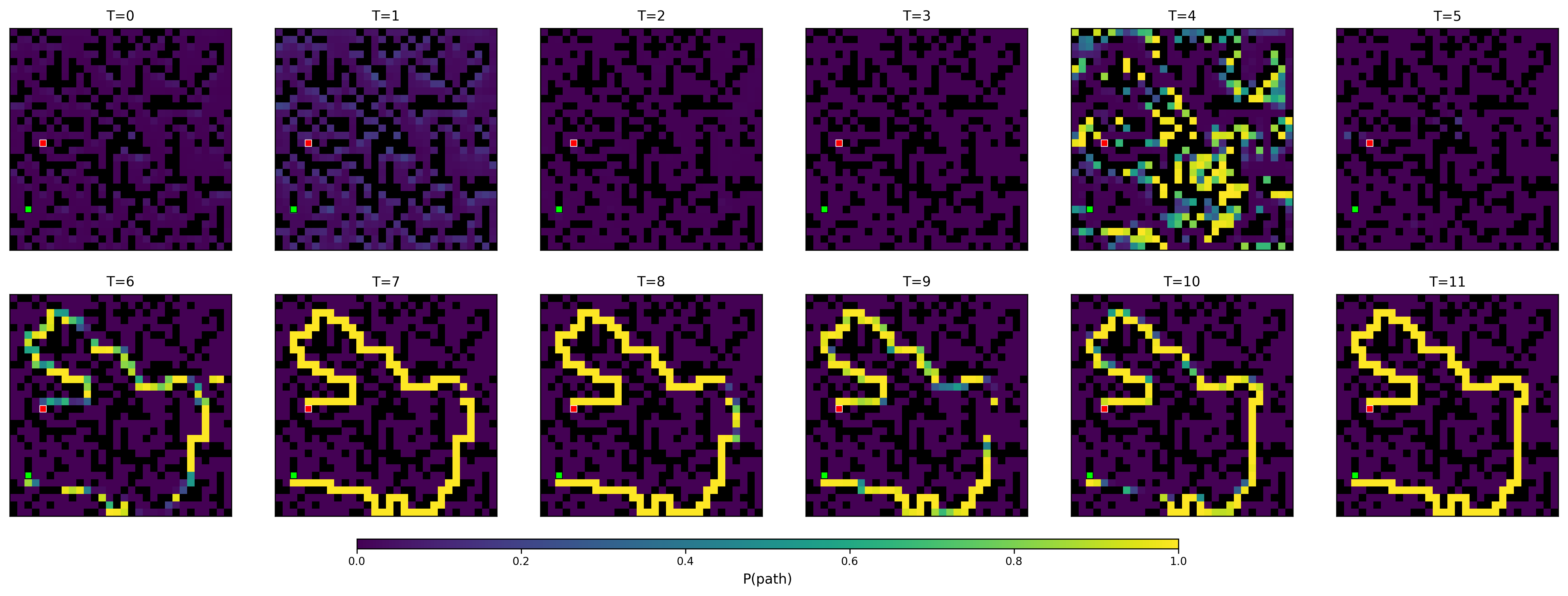}

    \caption{Partial temporal evolution of \emph{HRM} on the example shown in Sec.~\ref{sec:Maze_hard}.
    \textbf{Top:} discrete path predictions over time.
    \textbf{Bottom:} predicted path probability heatmaps over time.
    \emph{HRM} remains mostly inactive during early H-block updates and then undergoes an abrupt, insight-like transition, while \emph{WONN} progressively synchronizes diffuse candidate path fragments into a coherent valid path.
    }
    
    \label{fig:maze_hrm}
\end{figure}

Fig.~\ref{fig:maze_hrm} show the temporal evolution of \emph{HRM} on the same example discussed in Sec.~\ref{sec:Maze_hard}, including both path predictions and probability maps. We visualize the first 12 H-block updates and regard them as the first 12 evolution steps. The results suggest that \emph{HRM} remains largely inactive during the early stages, then begins to generate irregular predictions around $T=4$, and subsequently undergoes an abrupt transition around $T=6$ that recovers most of the correct path. The remaining steps only introduce minor refinements. This abrupt, insight-like behavior contrasts sharply with the progressive path formation exhibited by \emph{WONN}.

\newpage

\subsection{Additional Visualization on Sudoku}

\begin{figure}[t]
    \centering
    \includegraphics[width=\columnwidth]{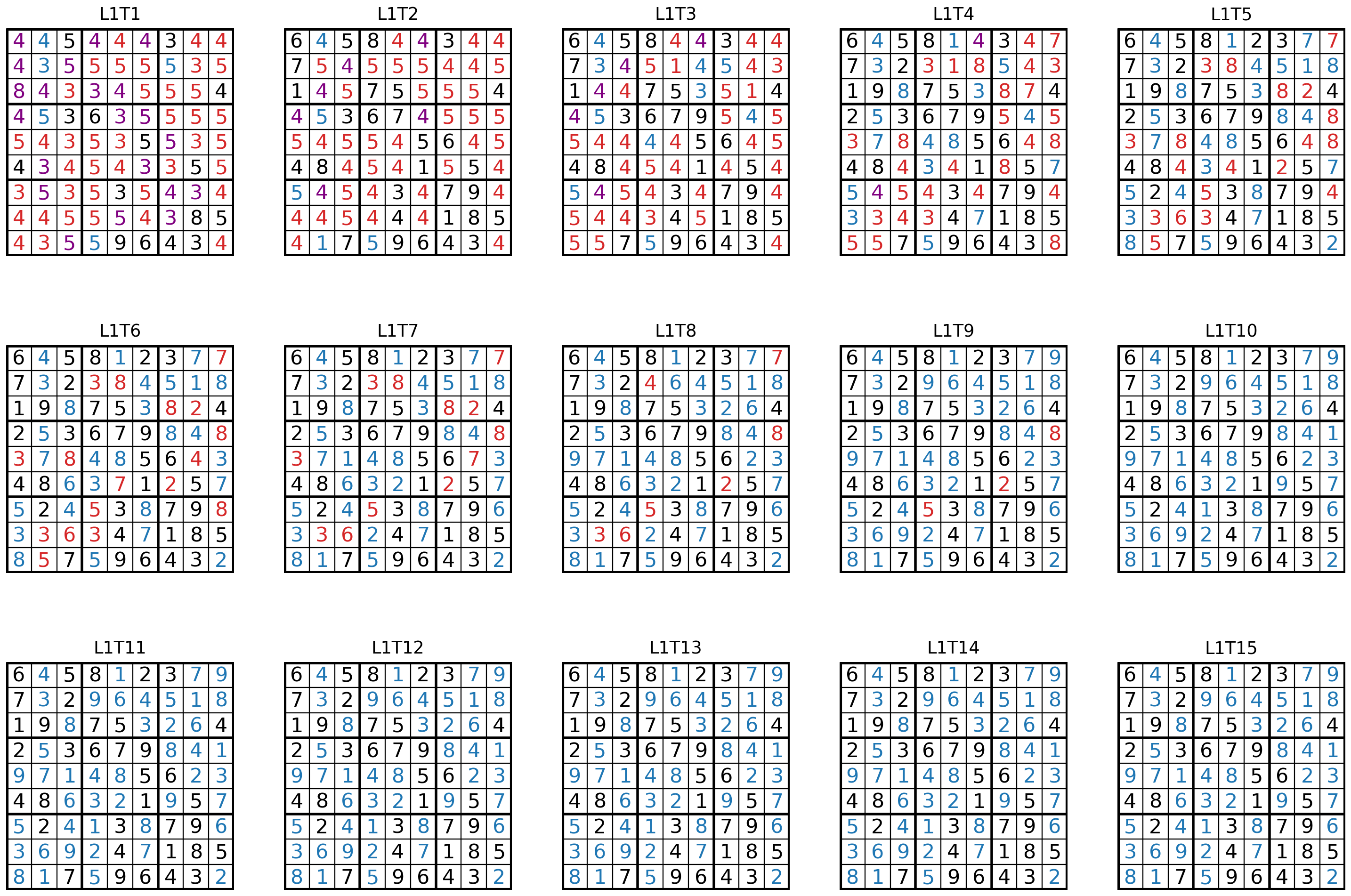}

    \vspace{5em}

    \includegraphics[width=\columnwidth]{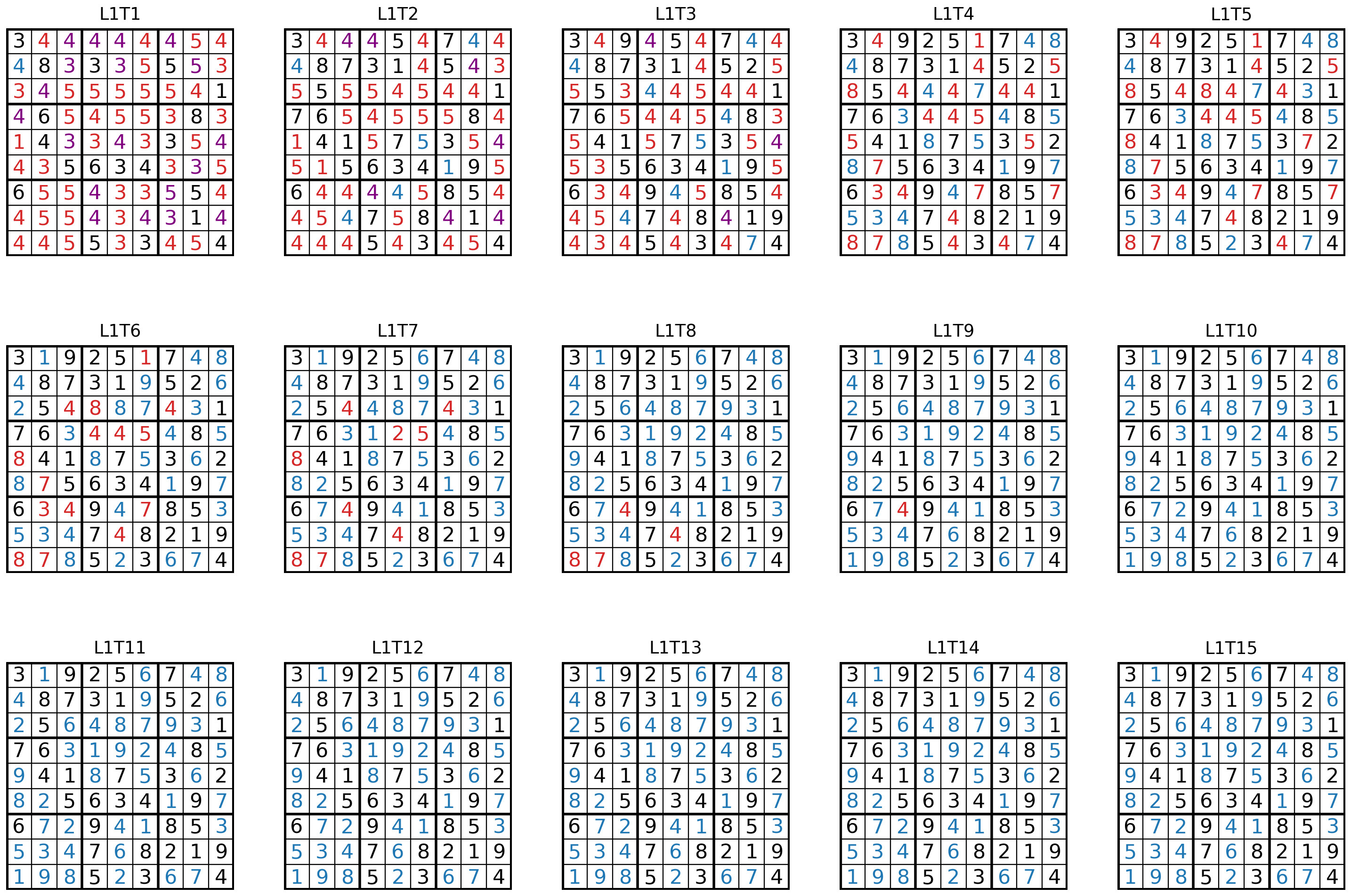}
    
    \caption{
    Temporal evolution of \emph{WONN} on Sudoku.
    Early steps show global synchronized exploration over candidate digit assignments, while later steps progressively refine these predictions and converge to a correct globally consistent solution.
    }
    \label{fig:sudoku_more_2}
\end{figure}

This section provides qualitative visualizations on the Sudoku task. Fig.~\ref{fig:sudoku_more_2} shows that \emph{WONN} follows a dynamical solving process similar to that observed on Maze-hard. At early time steps, the model tends to perform a global synchronized exploration of possible assignments. During subsequent evolution, these predictions are gradually refined, and the model progressively converges to the correct solution.

\FloatBarrier
\FloatBarrier

\FloatBarrier

\FloatBarrier
\begin{figure}[t]
    \centering

    \includegraphics[width=\columnwidth]{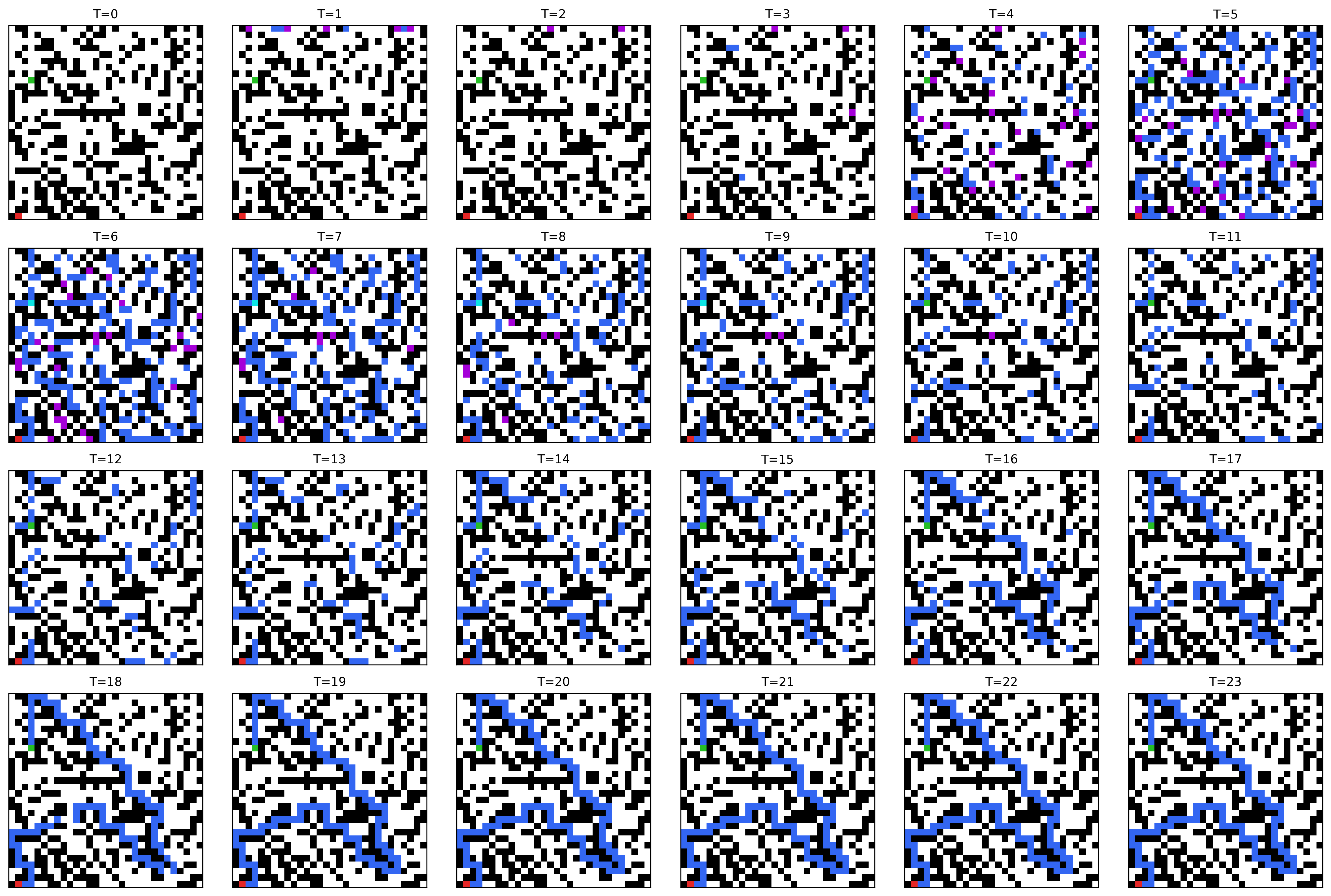}
    \vspace{0.5em}

    \includegraphics[width=\columnwidth]{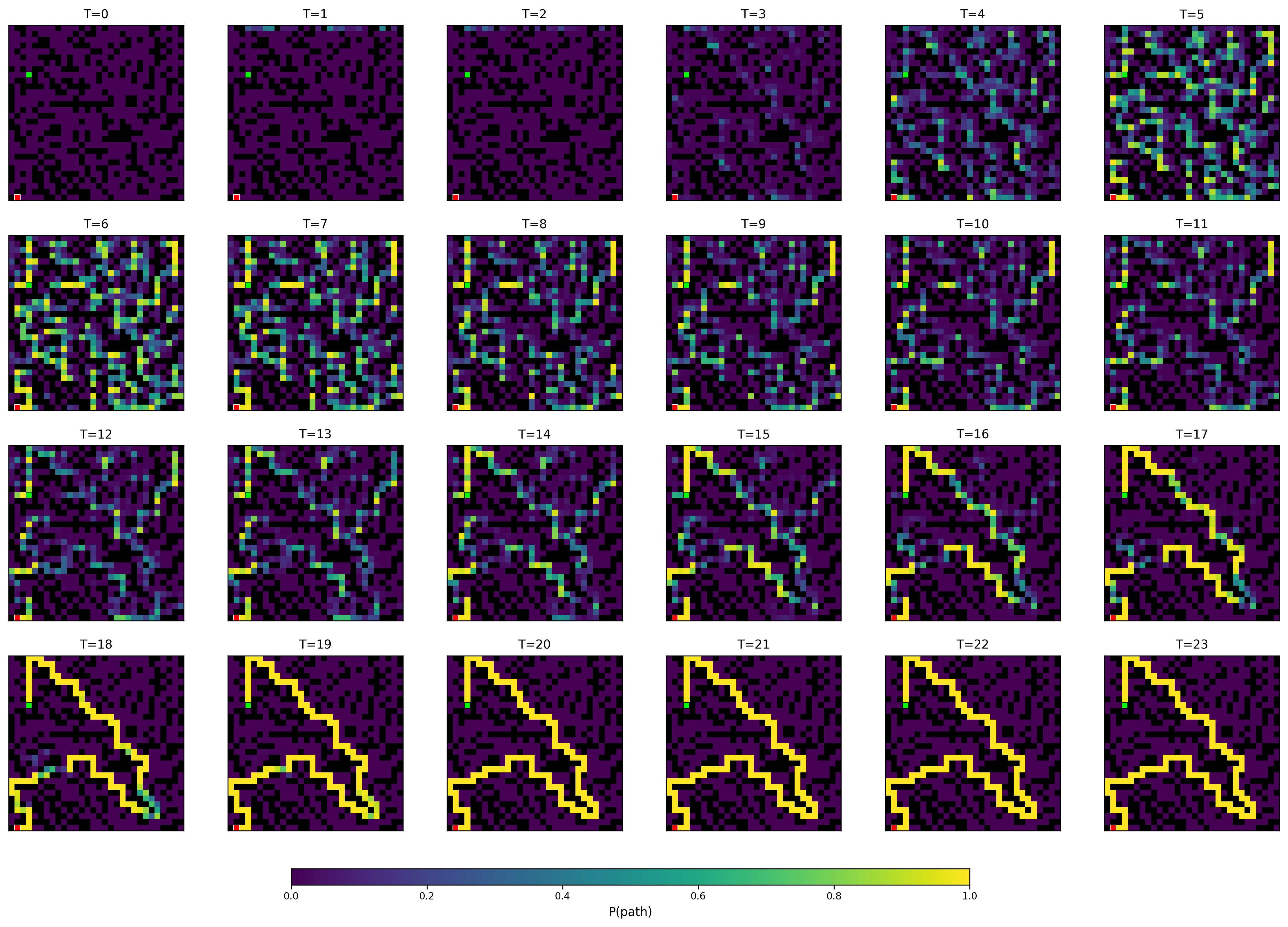}

    \label{fig:maze_path_more_2}
\end{figure}
\FloatBarrier

\FloatBarrier
\begin{figure}[t]
    \centering

    \includegraphics[width=\columnwidth]{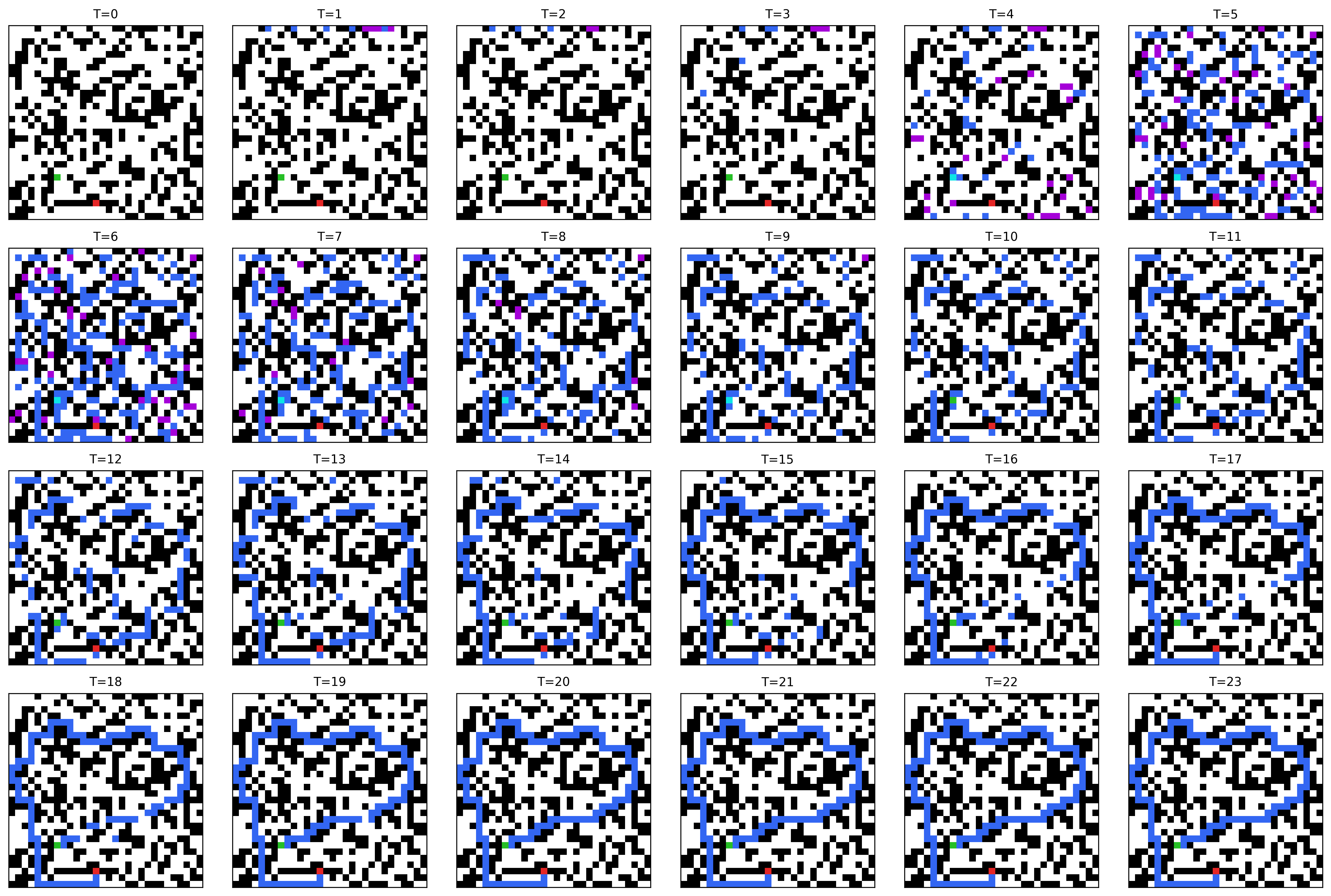}
    \vspace{0.5em}

    \includegraphics[width=\columnwidth]{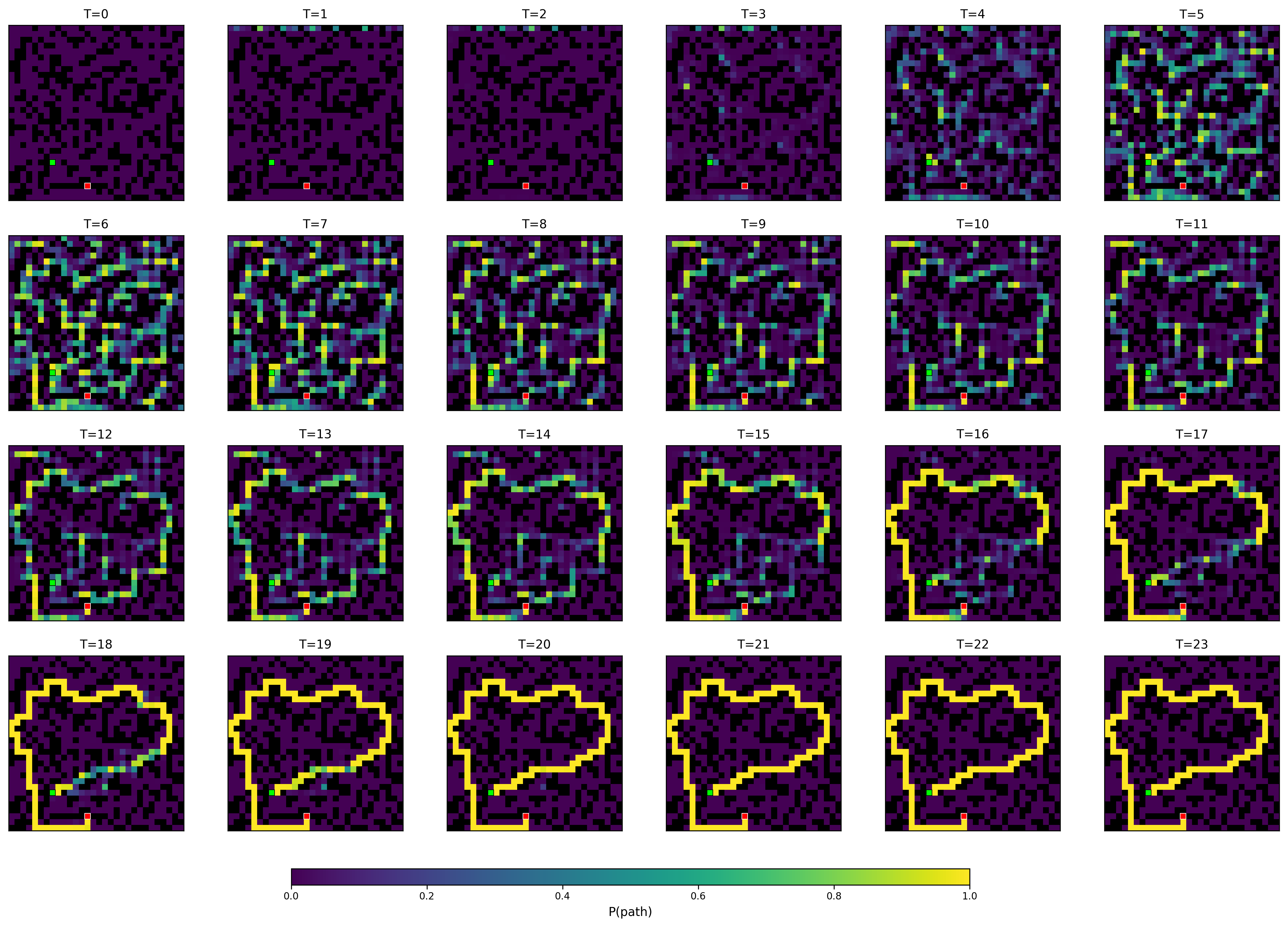}
    
    \label{fig:maze_path_more_3}
\end{figure}
\FloatBarrier

\FloatBarrier
\FloatBarrier

\FloatBarrier
\begin{figure}[t]
    \centering
    \includegraphics[width=\columnwidth]{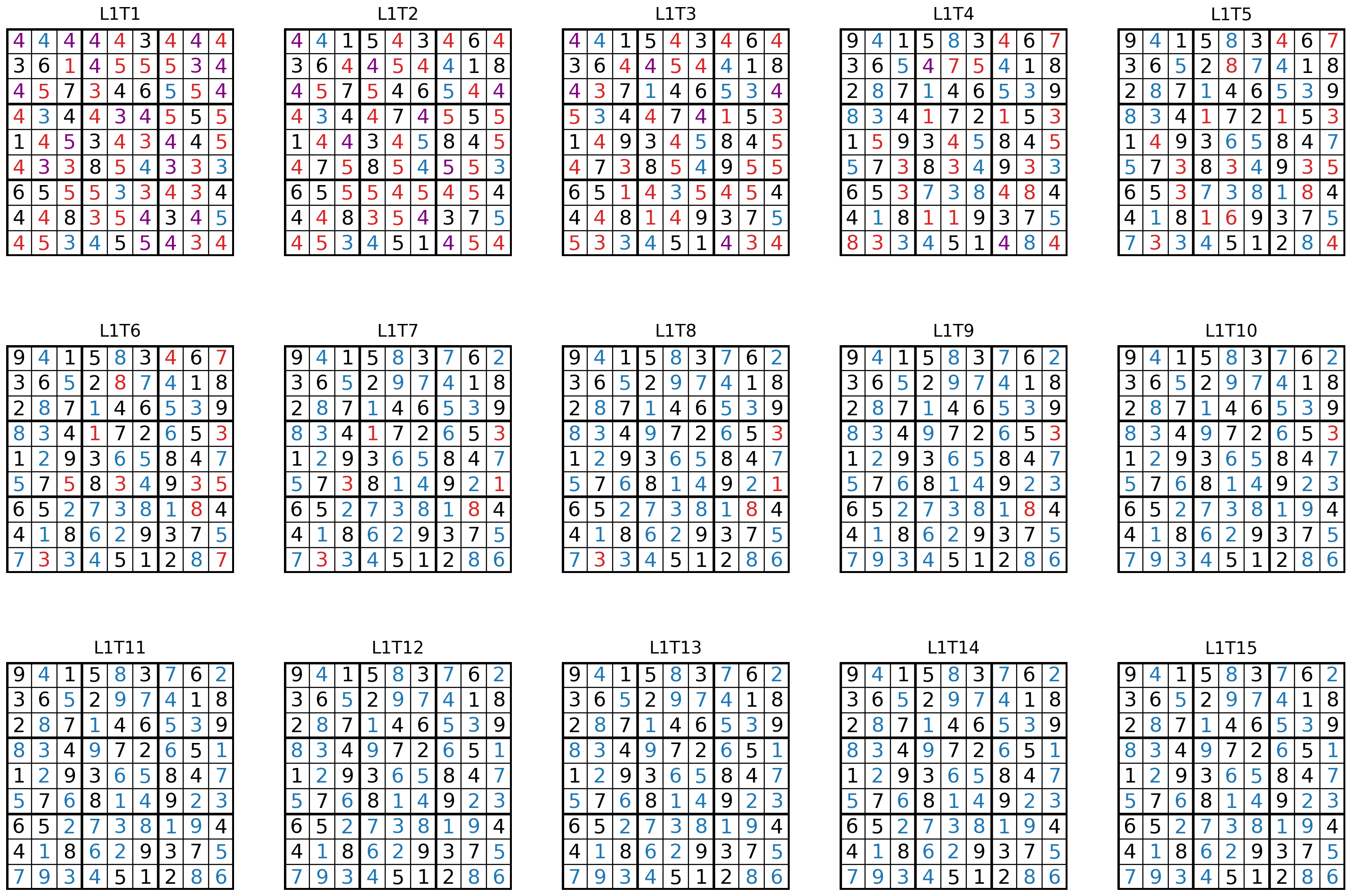}
    
    \vspace{5em}
    
    \includegraphics[width=\columnwidth]
    {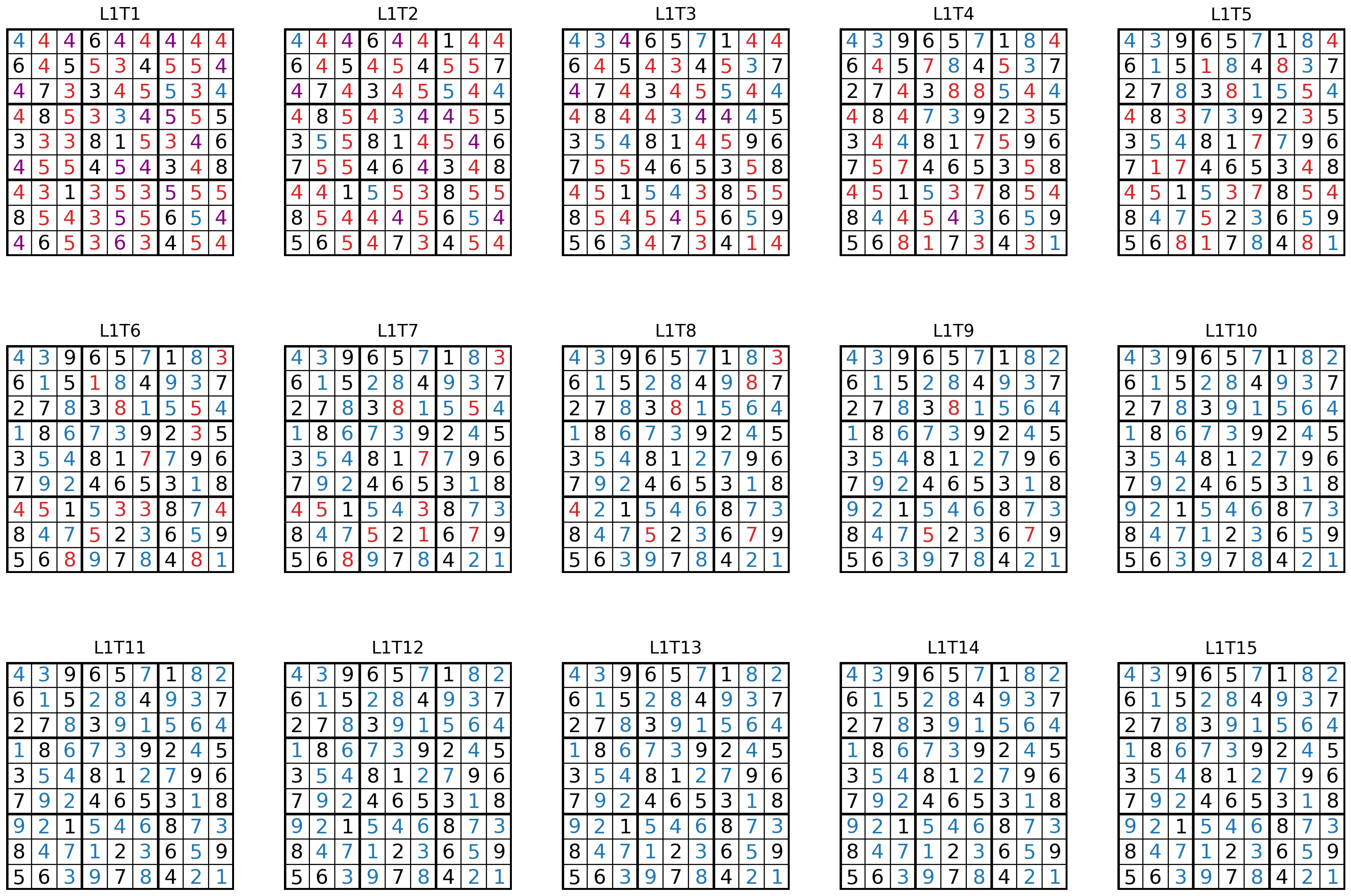}

    \label{fig:sudoku_more_1}
\end{figure}
\FloatBarrier

\FloatBarrier
\FloatBarrier

\newpage

\end{document}